\documentclass{article}
\usepackage[numbers,sort&compress]{natbib}

\usepackage[preprint]{neurips_2024}
\usepackage[utf8]{inputenc} % allow utf-8 input
\usepackage[T1]{fontenc}    % use 8-bit T1 fonts
\usepackage{hyperref}       % hyperlinks
\usepackage{url}            % simple URL typesetting
\usepackage{booktabs}       % professional-quality tables
\usepackage{amsfonts}       % blackboard math symbols
\usepackage{nicefrac}       % compact symbols for 1/2, etc.
\usepackage{microtype}      % microtypography
\usepackage{xcolor}         % colors
\usepackage{amsmath}
\usepackage{multirow}
\usepackage{arydshln}
\usepackage{tikz}
\usepackage{hyperref}

\usepackage{graphicx}
\usepackage{colortbl}
\usepackage{color}
\usepackage{subfigure}
\usepackage{diagbox}  % 提供 \diagbox 命令
\usepackage{array}    % 提供额外的表格功能

\usepackage{times}
\usepackage{textcomp}
\usepackage{enumitem}
\definecolor{yel}{RGB}{250,247,228}
\definecolor{gray}{RGB}{195,195,195}

\definecolor{world}{RGB}{55,158,49}
\definecolor{sports}{RGB}{254,131,2}
\definecolor{business}{RGB}{239,70,51}
\definecolor{scitech}{RGB}{98,64,150}

\usepackage{xspace}
\usepackage{pdfpages}

\newsavebox{\worldbox}  % 定义新的盒子
\savebox{\worldbox}{\tikz\draw[fill=world] (0,0) circle (.3em);}  % 在盒子中存储图形

\newsavebox{\sportsbox}  % 定义新的盒子
\savebox{\sportsbox}{\tikz\draw[fill=sports] (0,0) circle (.3em);}  % 在盒子中存储图形

\newsavebox{\businessbox}  % 定义新的盒子
\savebox{\businessbox}{\tikz\draw[fill=business] (0,0) circle (.3em);}  % 在盒子中存储图形

\newsavebox{\scitechbox}  % 定义新的盒子
\savebox{\scitechbox}{\tikz\draw[fill=scitech] (0,0) circle (.3em);}  % 在盒子中存储图形

\begin{document}

% \newsavebox{\mybox}  % 定义新的盒子
% \savebox{\mybox}{\tikz\draw[fill=world] (0,0) circle (.3em);}  % 在盒子中存储图形

\title{NoisyAG-News: A Benchmark for Addressing Instance-Dependent Noise in Text Classification}

% The \author macro works with any number of authors. There are two commands
% used to separate the names and addresses of multiple authors: \And and \AND.
%
% Using \And between authors leaves it to LaTeX to determine where to break the
% lines. Using \AND forces a line break at that point. So, if LaTeX puts 3 of 4
% authors names on the first line, and the last on the second line, try using
% \AND instead of \And before the third author name.

% \author{%
%   David S.~Hippocampus\thanks{Use footnote for providing further information
%     about author (webpage, alternative address)---\emph{not} for acknowledging
%     funding agencies.} \\
%   Department of Computer Science\\
%   Cranberry-Lemon University\\
%   Pittsburgh, PA 15213 \\
%   \texttt{hippo@cs.cranberry-lemon.edu} \\
%   % Address \\
%   % \texttt{email} \\
% }

\author{%
	Hongfei Huang, Tingting Liang, Xixi Sun, Zikang Jin, Yuyu Yin \\
	\texttt{\{h.fei.huang, liangtt, xixi.sun, 22320208, yinyuyu\}@hdu.edu.cn} \\
	Hangzhou Dianzi University 
}

\maketitle

\begin{abstract}
Existing research on learning with noisy labels predominantly focuses on synthetic label noise. Although synthetic noise possesses well-defined structural properties, it often fails to accurately replicate real-world noise patterns. In recent years, there has been a concerted effort to construct generalizable and controllable instance-dependent noise datasets for image classification, significantly advancing the development of noise-robust learning in this area. However, studies on noisy label learning for text classification remain scarce. To better understand label noise in real-world text classification settings, we constructed the benchmark dataset NoisyAG-News through manual annotation. Initially, we analyzed the annotated data to gather observations about real-world noise. We qualitatively and quantitatively demonstrated that real-world noisy labels adhere to instance-dependent patterns. Subsequently, we conducted comprehensive learning experiments on NoisyAG-News and its corresponding synthetic noise datasets using pre-trained language models and noise-handling techniques. Our findings reveal that while pre-trained models are resilient to synthetic noise, they struggle against instance-dependent noise, with samples of varying confusion levels showing inconsistent performance during training and testing. These real-world noise patterns pose new, significant challenges, prompting a reevaluation of noisy label handling methods. We hope that NoisyAG-News will facilitate the development and evaluation of future solutions for learning with noisy labels.
\end{abstract}

\section{Introduction}

% Extensive research has been conducted in the field of text classification, resulting in the introduction of widely adopted golden dataset benchmarks such as AG-News \cite{zhang2015character}, DBPedia \cite{auer2007dbpedia}, and TREC \cite{voorhees1999trec}. With the rapid advancement of deep learning techniques like CNNs, LSTMs, and Transformers, classification accuracy has seen significant improvements. The emergence of pre-trained language models (PLMs, e.g., BERT \cite{devlin2018bert}, XLNet \cite{yang2019xlnet}) has further substantially boosted task performance. Training these pre-trained models requires vast amounts of data \cite{marcelino2018transfer,han2021pre}, typically obtained through crowdsourcing or web crawling and subsequent annotation \cite{albert2022addressing}. However, owing to factors such as cognitive biases, varying annotator attentiveness \cite{peterson2019human}, or erroneous online information, datasets often contain real-world label noise, which can lead to overfitting on noisy labels and hinder the generalization capability of deep models. To mitigate this issue, numerous methods for robust deep model training under label noise have been extensively studied.

Extensive research in text classification has led to the development of widely adopted benchmark datasets such as AG-News \cite{zhang2015character}, DBPedia \cite{auer2007dbpedia}, and TREC \cite{voorhees1999trec}. With the rapid advancement of deep learning techniques like CNNs, LSTMs, and Transformers, classification accuracy has significantly improved. The emergence of Pre-Trained Language models (PLMs), such as BERT \cite{devlin2018bert}, XLNet \cite{yang2019xlnet}, has further boosted task performance. Training these models requires vast amounts of data \cite{marcelino2018transfer,han2021pre}, typically obtained through crowdsourcing or web crawling \cite{albert2022addressing}. However, factors such as cognitive biases, varying annotator attentiveness \cite{peterson2019human}, and erroneous online information often introduce real-world label noise, leading to overfitting and reduced generalization capability of deep models. 

% To address this issue, numerous methods for robust deep model training under label noise have been extensively studied.

% Substantial efforts have been dedicated to address the issue of label noise. Early research primarily focused on synthetic noise, particularly class-conditional noise (CCN) \cite{zhang2018generalized, lukasik2020does, ma2020normalized, cheng2022class}. However, synthetic noise, generated from predefined distributions, may not always effectively represent real-world conditions. Some studies \cite{tanzer2022memorisation,zhu2022bert} have shown that PLMs are robust to synthetic noisy labels but struggle with instance-dependent noisy labels. Recent research has recognized the prevalence of instance-dependent noise (IDN) and has explored it within the image domain \cite{liu2023identifiability, zhu2021second, chen2021beyond, yang2023parametrical, li2023disc, jiang2021information}. Despite these advancements, there is limited attention to the impact of instance noise on text classification tasks. Moreover, there is a lack of controlled text classification benchmark datasets to quantitatively evaluate and compare the effectiveness of different methods in handling instance-dependent noise.

Significant efforts have been made to address the issue of label noise. Early research primarily focused on synthetic noise, particularly class-conditional noise (CCN) \cite{zhang2018generalized, lukasik2020does, ma2020normalized, cheng2022class}. However, synthetic noise, generated from predefined distributions, may not always effectively represent real-world conditions. Some studies \cite{tanzer2022memorisation, zhu2022bert} have shown that PLMs are robust to synthetic noisy labels but struggle with instance-dependent noisy labels. Recent research has recognized the prevalence of instance-dependent noise (IDN) and has explored it within the image domain \cite{liu2023identifiability, zhu2021second, chen2021beyond, yang2023parametrical, li2023disc, jiang2021information}. Despite these advancements, limited attention has been given to the impact of instance noise on text classification tasks. Moreover, there is a lack of controlled text classification benchmark datasets to quantitatively evaluate and compare the effectiveness of different methods in handling IDN.

% Recent research has introduced several datasets containing real-world label noise. As illustrated in table \ref{tab:compareTable}, These general, real-world instance-dependent noise datasets have facilitated research on learning from noisy labels(LNL) methods in the field of CV. However, dataset benchmarks in the NLP domain lack generalizability and the ability to conduct controlled noise experiments for precise algorithm evaluation. NoisyNER \cite{hedderich2021analysing} focuses on identifying entities from sentences, which typically involve short samples and numerous word-level labels, making it challenging to determine the precise noise level of the dataset. NoisyWikiHow \cite{wu2023noisywikihow}, on the other hand, concentrates on intent analysis tasks and also suffers from the issue of relatively short sample sentences. Additionally, its specific characteristics, such as hierarchical categories and the presence of out-of-domain intent classes, limit its generalizability and make it difficult to ascertain whether the decline in classification accuracy when evaluating LNL methods stems from instance noise.Both datasets focus on specific tasks, where the texts to be classified and their categories lack generalizability. Additionally, they are not generated through human annotation, making it challenging to simulate real-world noise.

Recent research has introduced several datasets containing real-world label noise. As illustrated in Table \ref{tab:compareTable}, these general, real-world IDN datasets have facilitated research on learning from noisy labels (LNL) methods in the field of computer vision (CV). However, dataset benchmarks in the natural language processing (NLP) domain lack generalizability and the ability to conduct controlled noise experiments for precise algorithm evaluation. NoisyNER \cite{hedderich2021analysing} focuses on identifying entities from sentences, which typically involve short samples and numerous word-level labels, making it challenging to determine the precise noise level of the dataset. NoisyWikiHow \cite{wu2023noisywikihow}, on the other hand, concentrates on intent analysis tasks and also suffers from the issue of relatively short sample sentences. Additionally, its specific characteristics, such as hierarchical categories and the presence of out-of-domain intent classes, limit its generalizability and make it difficult to ascertain whether the decline in classification accuracy when evaluating LNL methods stems from instance noise. Both datasets focus on specific tasks, where the texts to be classified and their categories lack generalizability. Additionally, they are not generated through human annotation, making it challenging to simulate real-world noise.

\begin{table*}[ht]
\small%\setlength\tabcolsep{0.5pt}
\centering
\caption{
Comparison between Our Benchmark and Other Datasets. 
}
\begin{tabular}{lccccc}
\toprule
\textbf{Dataset}           & \textbf{Classes} & \textbf{Feature Size} & \textbf{General} & \textbf{\begin{tabular}[c]{@{}c@{}}Human\\ Annotation\end{tabular}} & \textbf{Size}  \\ \midrule
\rowcolor{yel} \multicolumn{6}{l}{CV}                                                                                \\
Food-101N~\citep{lee2018cleannet}         & 101  & Heterogeneous     & Yes    & No                                                    & 367K  \\
Animal-10N~\citep{song2019selfie}        & 10   & Heterogeneous   & Yes    & Yes                                                    & 55K   \\
CIFAR-10N ~\citep{wei2021learning}  & 10  & \(32 \times 32 \times3 \)    & Yes    & Yes                                                    & 55K   \\
CIFAR-100N~\citep{wei2021learning} & 100  & \(32 \times 32 \times3 \)     & Yes    & Yes                                                    & 16.1K \\
Clothing1M~\citep{xiao2015learning}        & 14   & Heterogeneous     & Yes    & No                                                    & 1M    \\ \midrule

\rowcolor{yel} \multicolumn{6}{l}{NLP}                                                                               \\
NoisyNER~\citep{hedderich2021analysing}          & 4    & 3    & No    & No                                                    & 14.8K \\ 
NoisywikiHow~\citep{wu2023noisywikihow}          & 158    & 10    & No    & No                                                    & 89K \\ \midrule
\textbf{NoisyAG-News}      & \textbf{4}  & \textbf{44}   & \textbf{Yes}    & \textbf{Yes}                                                    & \textbf{50K}   \\ \bottomrule
\end{tabular}
%\vspace{-0.1cm}

\label{tab:compareTable} 
%\vspace{-0.3cm}
\end{table*}

To bridge this gap, we introduce NoisyAG-News, a new text classification benchmark designed to evaluate the effectiveness of learning methods from instance-dependent label noise. The AG-News dataset is widely recognized and adopted for its ease of access and use. However, no publicly available non-expert human-annotated labels exist for the AG-News training dataset, hindering the evaluation of existing text classification methods and the validation of prevalent noise models. Therefore, we propose NoisyAG-News, a dataset with instance-dependent noise, constructed through non-expert, purely human crowdsourced annotation.

Due to the diverse backgrounds, preferences, biases, and annotation states of the annotators, conflicting labels may be assigned to the same ambiguous instance, resulting in instance-dependent label noise. Unlike previous efforts, such as NoisyNER \cite{hedderich2021analysing}, NoisyWikiHow \cite{wu2023noisywikihow}, and methods simulating instance noise \cite{xia2020part}, the NoisyAG-News dataset is more general and realistic, serving as the first controlled text classification benchmark that includes real-world noise. This will significantly enhance the evaluation of current and future solutions for handling instance-dependent label noise in text classification tasks. Our contributions can be summarized as follows:

\begin{itemize}
\item[$\bullet$] We introduce a new benchmark dataset, NoisyAG-News, derived from crowdsourced redundant label sets. This dataset will be maintained to support future research on addressing IDN in text classification tasks.
\item[$\bullet$] We investigated the characteristics of noise transition matrices derived from manual annotations and sets of samples with varying degrees of confusion. Both qualitative and quantitative analyses were conducted to differentiate between manually annotated noise in NoisyAG-News and synthetic noise, substantiating that the noise in our dataset is instance-dependent.
\item[$\bullet$] We compared the effectiveness and stability of various PLMs and LNL methods on different noisy datasets. The results indicate that IDN are more difficult to distinguish compared to synthetic noise. This highlights a significant challenge and underscores the importance of our proposed benchmark.
\end{itemize}
% \section{Preliminary}

\section{Noise Type}
\label{sec:NOiseType}
suppose there is a training dataset  \( D:=\left\{\left(x_{n}, y_{n}\right)\right\}_{n \in[N]} \), where  \( [N]:=\{1,2, \ldots, N\} \) . In real-world scenarios, a classifer \( f \) only has access to noisily labeled training set \( \widetilde{D}:=\left\{\left(x_{n}, \tilde{y}_{n}\right)\right\}_{n \in[N]} \). compare  \( D \) with \( \widetilde{D} \)
, It's evident that there may exist \( {n \in[N]} \) such that \( y_{n} \neq \tilde{y}_{n} \) , The ﬂipping from clean to noisy label is usually formulated by a Noise Transition Matrice  (NTM)  \( T(X) \), with elements: \( T_{i, j}(X):=\mathbb{P}(\widetilde{Y}=j \mid Y=i, X) \) , where \(X\) represents features, and \(Y \)  represents labels. we shall specify  \( T(x) \)  for different noise below.

{\bf Uniform Noise }  
It assumes that the probability of randomly ﬂipping the clean class to the other possible class with probability \( \epsilon \). Assuming a noise level of \( \epsilon \) , the diagonal entry of the symmetric T is denoted as \( T_{i, i}=1-\epsilon \). For any other off-diagonal entry \( T_{i, j} \) where \(i \neq j \), the corresponding element is \( T_{i, j}=\frac{\epsilon}{K-1} \) \cite{ma2018dimensionality,jenni2018deep,jindal2016learning,yuan2018iterative}.

{\bf Single-Flip Noise }   
Single-Flip Noise is assumed to be conditionally independent of the feature $X$. Mathematically,$T(X)=T$ and \(T_{i, j}(X)=\mathbb{P}(\tilde{Y}=j \mid Y=i),  \forall i, j \in[K] \) , Single-Flip NTMs assumes that the clean label ﬂips to the next class with probability $\epsilon$, i.e, $i \rightarrow(i+1) \bmod K$ for $i \in[K]$ \cite{han2018co,ren2018learning,yu2019does}.

{\bf Synthesized Instance-Dependent Noise }  
Recent studies \cite{chen2021beyond,zhao2022centrality,wang2021tackling} \cite{xia2020part,zhu2021second,yang2023parametrical,cheng2020learning,liu2023identifiability,yao2021instance,cheng2022instance,garg2023instance,berthon2021confidence} have recognized the presence of instance-dependent noise, One of the most widely used approaches\cite{xia2020part} involves multiplying sample features by a common variable \( W \). This method leverages the similarity of sample features to flip true labels to noisy labels according to a specific noise rate. 

{\bf Flip Noise by NTM} 
After setting a noise rate \( \epsilon \), the noise rates for each class in datasets synthesized using Uniform Noise, Single-Flip Noise, and Synthesized Instance-Dependent Noise are all equal to \( \epsilon \). However, in real-world scenarios, the noise rates for samples from different classes are not consistent. To better compare the instance-dependent noise derived from annotations with class-conditional noise, we randomly flipped the GT labels to noisy labels according to the NTM corresponding to NoisyAG-News.

{\bf Instance-Dependent Noise }
Beyond the feature-independent assumption \cite{chen2021beyond}, recent works pay more attention to a challenging case where the label noise is jointly determined by feature \( X \) and clean label \( Y \) ,denoted by  \( T_{i, j}(X)=\mathbb{P}(\tilde{Y}=j \mid Y=i,X = x), \forall i, j \in[K] \) . 

% \subsection{Learn from Noise}

% Numerous methods have been developed to address label noise in image classification, such as selecting potentially clean samples for training \cite{yao2020searching, han2020sigua, han2018co, li2020dividemix, chang2023csot}, adding regularization \cite{li2020gradient, guo2018curriculumnet, li2017learning, han2018masking, wu2021class2simi}, and loss adjustment \cite{zhang2018generalized, amid2019robust}. In contrast, only a few studies \cite{jindal2019effective, zheng2021meta, garg2021towards, lee2022context} have focused on label noise in text classification, and their experiments have been conducted on synthetic noisy datasets.
\section{NoisyAG-News: Human Annotated Noisy Labels on AG-News}

\subsection{Annotation Process}

We selected 50,000 samples for manual annotation from the AG-News dataset, with 12,500 samples per category. To obtain crowdsourced redundant annotations, 60 annotators were divided into three groups, each annotating the entire dataset, resulting in three labels per sample. A preliminary test was conducted on 4,000 samples, with each annotator responsible for 200 samples. Annotations were reviewed for quality, and inter-annotator agreement was assessed using Cohen’s Kappa, yielding a score of 0.75. After validation, the remaining 46,000 samples were annotated similarly, providing a complete set of annotated labels. Detailed descriptions and analysis are provided in Appendix \ref{supplement:A}. 

\subsection{Label Aggregation}

Based on the three annotated labels assigned to each sample, three datasets with varying levels of label noise were constructed: NoisyAG-NewsBest, NoisyAG-NewsMed and NoisyAG-NewsWorst, with noise ratios incrementally increasing from low to high, respectively. The procedures for obtaining sample labels in these noise-induced datasets are detailed as follows:

\begin{itemize}
\item[$\bullet$] {\bf NoisyAG-NewsBest: } When one of the annotation labels matches the true label, the sample label is set to the true label. Otherwise, one of the annotation labels is randomly selected, resulting in a dataset with an accuracy of approximately 90\% and a noise rate of about 10\%.
\item[$\bullet$] {\bf NoisyAG-NewsMed:} Labels are aggregated using a majority voting approach to generate the sample label, resulting in a dataset with an accuracy of approximately 80\% and a noise rate of about 20\%.
\item[$\bullet$] {\bf NoisyAG-NewsWorst:} When any of the annotated labels differ from the true label, the sample label is randomly selected from the differing labels. Otherwise, it is set to the true label. This method constructs a dataset with an accuracy of approximately 62\%, corresponding to a noise rate of about 38\%.
\end{itemize}

\begin{figure}
	\centering
	\includegraphics[width=12cm]{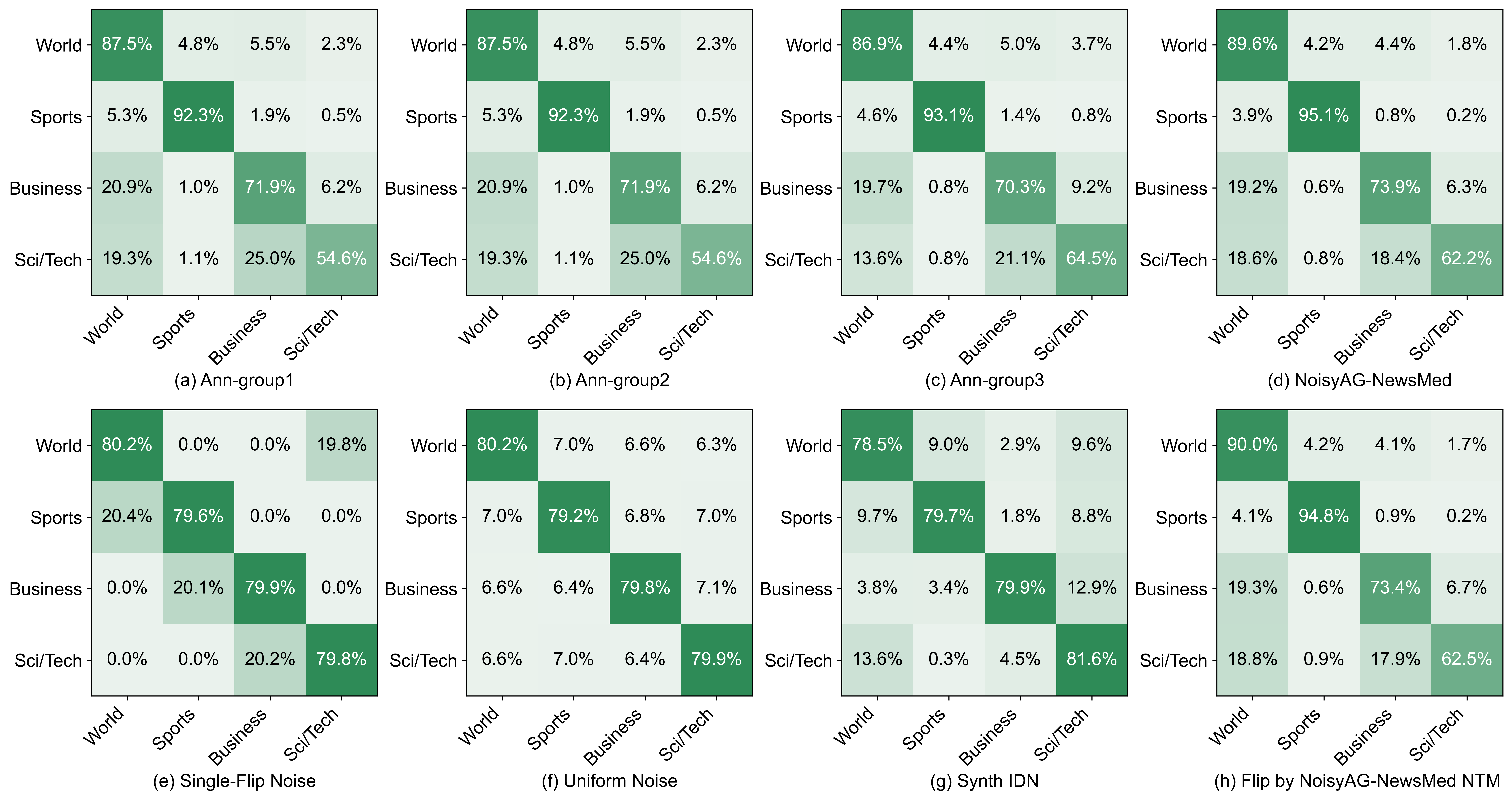}
	\caption{NTMs for Three Annotation Group, NoisyAg-NewsMed and Corresponding Synthetic Noise. }
        \label{fig:matShow} 
\end{figure}

% Based on the true labels of the samples and the three redundant annotation labels, we construct two datasets with different noise rates: NoisyAG-NewsBest and NoisyAG-NewsWorst. For controversial samples, NoisyAG-NewsBest was assigned the true label, while NoisyAG-NewsWorst was assigned a noise label. NoisyAG-NewsMed, on the other hand, did not rely on true labels; it was constructed solely based on the annotation labels, consistent with real-world crowdsourcing scenarios. The accuracy and noise rates of the three datasets were calculated using the true labels, allowing us to conduct experiments under controlled conditions. 

\subsection{Pattern of NTMs}

As depicted  by Figure \ref{fig:matShow} (a) - (c), we plotted the NTMs corresponding to the three annotated datasets. Panel (d) presents the NTM for NoisyAG-NewsMed, while panels (e) to (h) illustrate the NTMs for various synthetic noises. Compared to the NTMs of synthetic noise, the NTMs of the NoisyAG-News datasets appear disorderly and lack distinct patterns. However, a similar pattern emerged: clean labels were more frequently flipped to one or more similar classes. The Science/Technology class was prone to being misclassified as World and Business, while the Sports class demonstrated higher precision and recall compared to other classes, suggesting that sample flipping might be feature-dependent. More comparsions of the NoisyAG-News can be found in Appendix \ref{supplement:B}.

\subsection{Real-world Noise vs. Synthetic Noise}
\subsubsection{A Qualitative Aspect}
\label{sec:Qualitative}  

According to \cite{wei2021learning}, we have the definition of M-NN noise clusterability, and 
\( T(X) \) can be estimated on the set of neighboring samples \( \widetilde{\mathcal{D}}_{n}\). Utilizing the KNN algorithm, the features \( X \) were automatically clustered into \( K \) clusters, thus determining M as the number of samples in each cluster. Let \( \mathcal{I}_{i, \nu} \) denote the set of instance indices from the \( \nu \) -th cluster of clean class \( i \). Then, for label noise in \( \mathcal{I}_{i, \nu} \), it was assumed to be feature-independent, and the corresponding transition vector \( {p}_{i, \nu} \) was defined, where each element \( {p}_{i, \nu}[j] \) is expected to be \( \mathbb{P}(\tilde{Y}=j \mid x_{n}, n \in \mathcal{I}_{i, \nu}, Y=i)\). The transition vector \({p}_{i, \nu} \) was estimated by counting the frequency of each noisy class given noisy labels in \(\mathcal{I}_{i, \nu}\) .

For comparative analysis, the transition vectors for the World and Sci/Tech classes in NoisyAG-NewsMed and the corresponding synthetic noise datasets are presented in Figure \ref{fig:matTrans}. It can be observed that in Figure \ref{fig:matTrans}, the rows of each matrix for synthetic noise are very similar, indicating that the synthetic noise is feature-independent. In contrast, the transition vectors for different feature clusters of the World and Sci/Tech classes in NoisyAG-News are significantly different, suggesting that the noise transformations \( p_{i, \nu}[j] \) depend on features. This qualitatively demonstrates that the noise in NoisyAG-News is instance-dependent. Further comparisons of more classes, as well as NoisyAG-NewsBest, NoisyAG-NewsWorst, and synthetic noise,  can be found in Appendix \ref{supplement:D}.

\subsubsection{A Quantitative Aspect}

\begin{figure}
	\centering
	\includegraphics[width=12cm, trim={0cm 0cm 0cm 0cm}, clip]{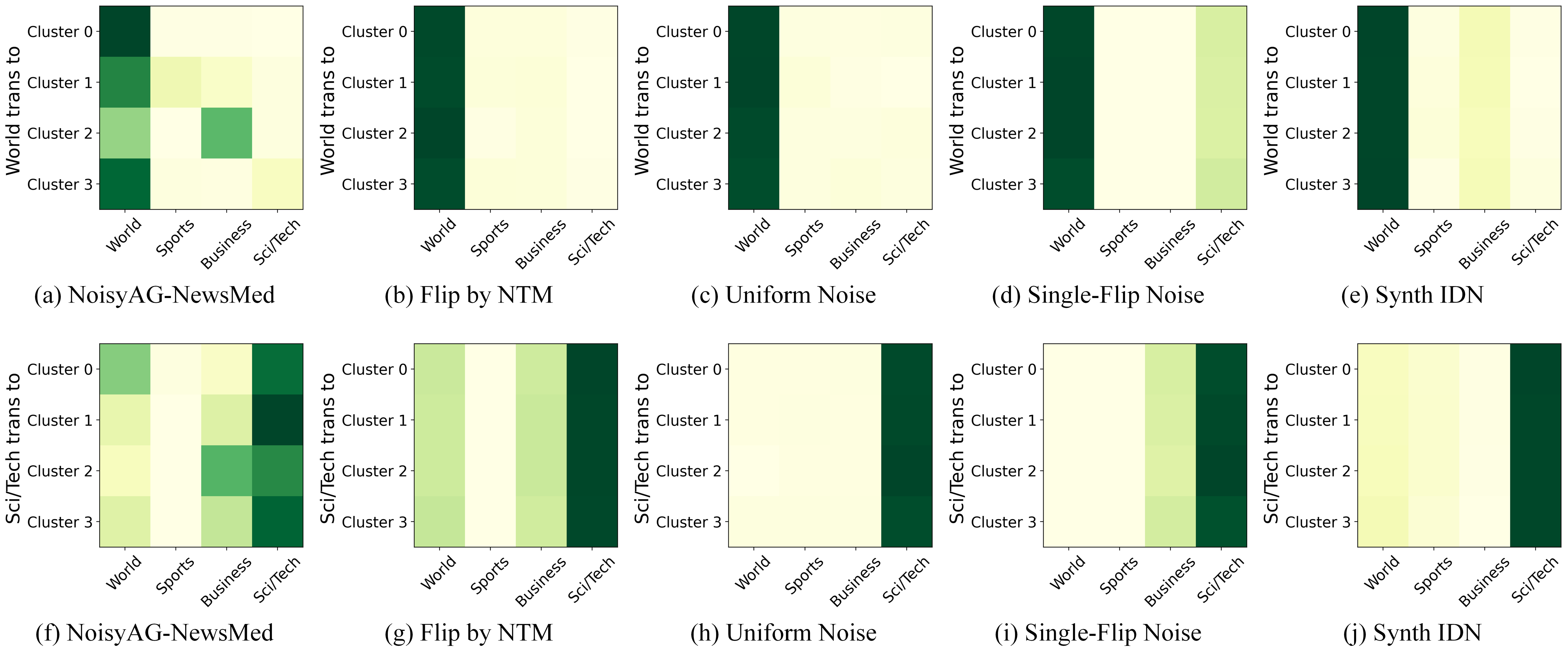}
	\caption{ \( {p}_{i, \nu}[j]\) in NoisyAG-NewsMed and Synthetic Noise for Class World and Sci/Tech.}
        \label{fig:matTrans} 
\end{figure}

Based on the clustering results from the previous section \ref{sec:Qualitative} , we further statistically test whether human noise is dependent on features. The null hypothesis \( H_0 \) and the corresponding alternative hypothesis \( H_1 \) are defined as:
\begin{equation*}
    H_0: \text{Human label noise is feature-independent.}
\end{equation*}
\begin{equation*}
    H_1: \text{Human label noise is feature-dependent.}
\end{equation*}
As evidenced by the mathematical principles of the synthesized noise and the visualization in Figure \ref{fig:matTrans}, it is feature-independent.One measure of the discrepancy between human noise and synthetic noise is the distance between their respective transition vectors  \( {p}_{i, \nu} \) across different noise clusters. For a comparative analysis, we need to compare the following:

\[ d_{i, \nu}^{(1)}:=\left\|\boldsymbol{p}_{i, \nu}^{\text {human }}-\boldsymbol{p}_{i, \nu}^{\text {synthetic }^{'}}\right\|_{2}^{2} \quad , \quad
 d_{i, \nu}^{(2)}:=\left\|\boldsymbol{p}_{i, \nu}^{\text {synthetic }}-\boldsymbol{p}_{i, \nu}^{\text {synthetic }^{'}}\right\|_{2}^{2}. \]

Let \( \boldsymbol{p}_{i, \nu}^{\text{synthetic }} \)  and  \( \boldsymbol{p}_{i, \nu}^{\text{synthetic }^{'}} \) represent the transition matrices of two synthesized noisy label datasets, while \( \boldsymbol{p}_{i, \nu}^{\text {human }}  \)  denotes the transition matrix of NoisyAG-News. By fixing  \( \boldsymbol{p}_{i, \nu}^{\text{synthetic }} \)  and \( \boldsymbol{p}_{i, \nu}^{\text {human }}  \) , and synthesizing multiple instances of \( \boldsymbol{p}_{i, \nu}^{\text{synthetic }^{'}} \) , the distances represented by the aforementioned two formulas were calculated to characterize the discrepancy between  \( \boldsymbol{p}_{i, \nu}^{\text{synthetic }} \)  and \( \boldsymbol{p}_{i, \nu}^{\text {human }}  \) .
To quantify this comparison, we employ a statistical hypothesis test and perform a two-sample t-test to determine if the mean distances under the two conditions are significantly different. Let \( h \) and \( s \) denote the mean distances under human and synthesized noise respectively. The null and alternative hypotheses for the t-test can be formulated as:
\begin{equation*}
    H_0: h = s \quad \text{(Human noise is feature-independent, same as synthesized noise) .}
\end{equation*}
\begin{equation*}
    H_1: h > s \quad \text{(Human noise is feature-dependent, with larger distances than synthesized noise).}
\end{equation*}

We calculated the average distances \(h \) and  \( s \) by synthesizing ten instances of  \( \boldsymbol{p}_{i, \nu}^{\text{synthetic }^{'}} \) ,  Based on these average distances, we performed a statistical significance test as shown in Table \ref{tab:Quantitative}.

\begin{table}[htbp]
\small 
\centering
\caption{ \( p\)-value between Ours Benchmark and Different Synthetic Noise.   }
\label{tab:Quantitative}
\scalebox{0.8}{
% 增加行间距，调整为1.5或根据需要选择合适的数值
\renewcommand{\arraystretch}{1.2}
\begin{tabular}{ccccc}
\toprule
\multirow{2}{*}{Benchmark} & \multicolumn{4}{c}{Noise type} \\
\cline{2-5}
& Flip by NTM & Uniform & Single-Flip & Synth-IDN \\
\toprule
NoisyAG-NewsBest  & \(2.1\times 10^{-34}\) & \( 6.8\times 10^{-36} \) & \( 4.39\times 10^{-37}\) & \(  7.1\times 10^{-19} \)  \\
\midrule
NoisyAG-NewsMed  & \(4.3\times 10^{-36}\)  & \(  6.3\times 10^{-40} \) & \( 4.13\times 10^{-39}\) & \( 5.0\times 10^{-17}\)   \\
\midrule
NoisyAG-NewsWorst  & \(2.2\times 10^{-36}\) & \( 4.8\times 10^{-41} \) & \( 1.3\times 10^{-44}\) & \(4.1\times 10^{-11}\) \\
\bottomrule
\end{tabular}}
\end{table}

The extremely small p-values allow us to firmly reject the null hypothesis \( H_0 \) and support the alternative hypothesis \( H_1 \). Consequently, through this quantitative analysis, we can conclude that the human-annotated noisy label dataset NoisyAG-News significantly differs from the synthesized class-conditional noisy label dataset, suggesting that the noise in NoisyAG-News is instance-dependent rather than solely relying on class information. More information regarding the qualitative and quantitative distinctions between our benchmark and synthetic noise will be provided in Appendix \ref{supplement:D}.

\section{Experiment and Discussion}

\subsection{Experiment Setting}

{\bf Training Setting } 
We utilized a server equipped with an Intel Platinum 8358 CPU, 196GB of memory, and A6000 Ada GPUs for our experiments. The training steps was set to 20,000. The detailed experimental setup is shown in Appendix \ref{supplement:E}.

% Validation and testing were performed every 100 steps. Additionally, we implemented an early stopping technique, halting training if the loss did not show significant improvement over 36 consecutive steps.

\begin{figure}
	\centering
	\includegraphics[width=12cm]{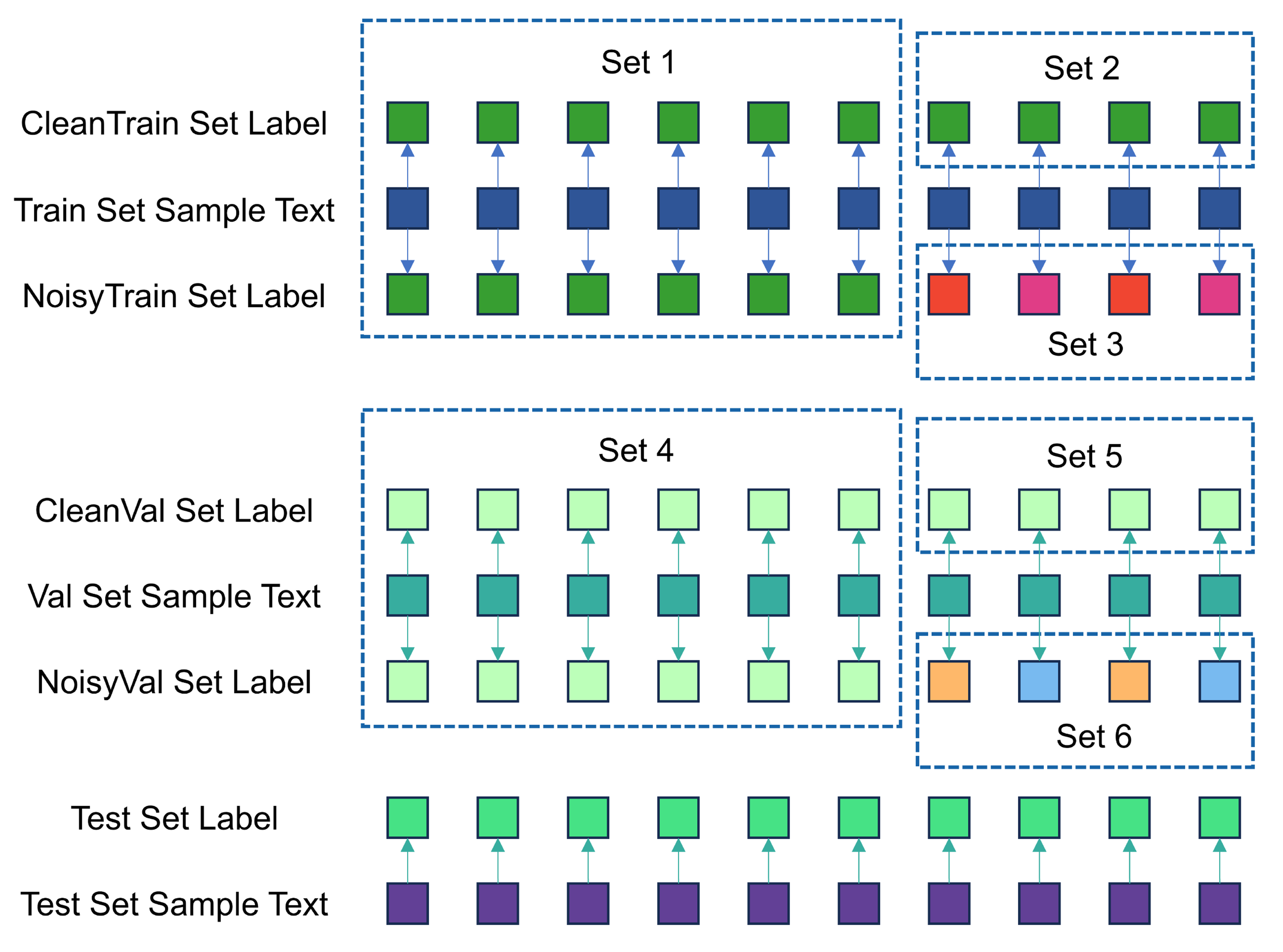}
	\caption{DataSet Decomposition, State Definition and Transition.}
        \label{fig:dataSplitAndState}
\end{figure}

{\bf DataSet Decomposition and Label Flip } To better describe the learning process of the neural network and analyze the impact of different parts of the dataset, we explain how to generate the training set, validation set, test set, and noisy data subsets. Firstly, the 50,000 samples are shuffled, with 90\% selected as the training set and 10\% as the validation set. In generating noise, a specific noise ratio is applied to select a corresponding number of samples from the training set. As illustrated in Figure \ref{fig:dataSplitAndState} (a), the ground truth labels of these selected samples form Set2. The labels in Set2 are then flipped to noisy labels, resulting in Set3. The same procedure is used to create Set5 and Set6. The test set comprises the original 7,600 samples.

{\bf Synthetic Noise} We synthesized four distinct types of noise as mentioned in \ref{sec:NOiseType}: flipping to noise labels by NoisyAG-News NTM, Single-Flip noise, Uniform noise, and synthetic IDN. We refer to these synthesized noisy datasets as follows: NoisyAG-News-NTM, NoisyAG-News-SingleFlip, NoisyAG-News-Uniform, and NoisyAG-News-SynthIDN.

\subsection{Performance Comparisions}

We conducted experiments using different models \cite{devlin2018bert,liu2019roberta,he2021debertav3,lewis2019bart,yang2019xlnet}, as depicted in Table\ref{tab:Performance}, When evaluating model performance across the NoisyAG-News dataset and its synthetic noise counterparts, significant differences were observed. On the NoisyAG-News dataset, a notable decline in classifier accuracy highlighted the difficulties current models face when handling real, feature-dependent label noise in text classification tasks. Unlike the synthetic noise datasets, the NoisyAG-News dataset is entirely annotated by humans, capturing authentic noise patterns. This distinction makes it a more challenging and practically relevant benchmark for assessing the robustness of models to noise.

\begin{table*}[]
\centering
\caption{Performance of Various Models Under Different Noise Levels.}
\scalebox{0.72}{
\begin{tabular}{@{}llllllllllllllllll@{}}
\toprule
 & \multicolumn{1}{c}{} & \multicolumn{3}{c}{Ours} & \multicolumn{3}{c}{Flip by NTM} &  \multicolumn{3}{c}{Synth IDN}  &  \multicolumn{3}{c}{Single-Flip}  &  \multicolumn{3}{c}{Uniform} 
 \\ \cmidrule(lr){3-5} \cmidrule(lr){6-8} \cmidrule(lr){9-11}  \cmidrule(lr){12-14} \cmidrule(lr){15-17} 
  & \multicolumn{1}{c}{Clean} & \multicolumn{1}{c}{10\%} & \multicolumn{1}{c}{20\%} & \multicolumn{1}{c}{38\%} & \multicolumn{1}{c}{10\%} & \multicolumn{1}{c}{20\%} & \multicolumn{1}{c}{38\%}  & \multicolumn{1}{c}{10\%} & \multicolumn{1}{c}{20\%} & \multicolumn{1}{c}{38\%} & \multicolumn{1}{c}{10\%} & \multicolumn{1}{c}{20\%} & \multicolumn{1}{c}{38\%} & \multicolumn{1}{c}{10\%} & \multicolumn{1}{c}{20\%} & \multicolumn{1}{c}{38\%} \\ \midrule

BERT & 93.4 & 90.0 & 85.8 & 77.4 & 92.9 & 92.4 & 87.9 & 92.9 & 92.2 & 91.0 & 92.9 & 92.9 & 91.7 & 92.7 & 92.4 & 91.6  \\
RoBERTa & 93.9 & 90.6 & 86.9 & 77.1 & 93.3 & 92.7 & 90.2 & 93.4 & 92.8 & 92.2 & 93.5 & 93.4 & 93.1 & 93.3 & 93.0 & 91.9 \\
DeBERTa-V3 & 93.6 & 90.3 & 86.3 & 77.7 & 93.3 & 92.8 & 90.5 & 93.2 & 92.9 & 92.0 & 93.3 & 93.0 & 92.8 & 93.2 & 93.4 & 92.0 \\
BART & 93.7 & 90.4 & 86.8 & 78.5 & 93.3 & 92.6 & 90.0& 93.2 & 92.6 & 92.0 & 93.5 & 93.0 & 92.0 & 93.2 & 92.9 & 92.0 \\
XLNET & 93.8 & 90.2 & 86.6 & 77.4 & 92.8 & 92.3 & 86.8 & 92.6 & 92.1 & 90.9 & 93.0 & 92.7 & 91.6 & 92.6 & 92.9 & 91.5 \\
 \bottomrule
\end{tabular}}

\label{tab:Performance}
\end{table*}

\subsection{Acc on Different Dataset}

\begin{figure}
	\centering
	\includegraphics[width=12cm]{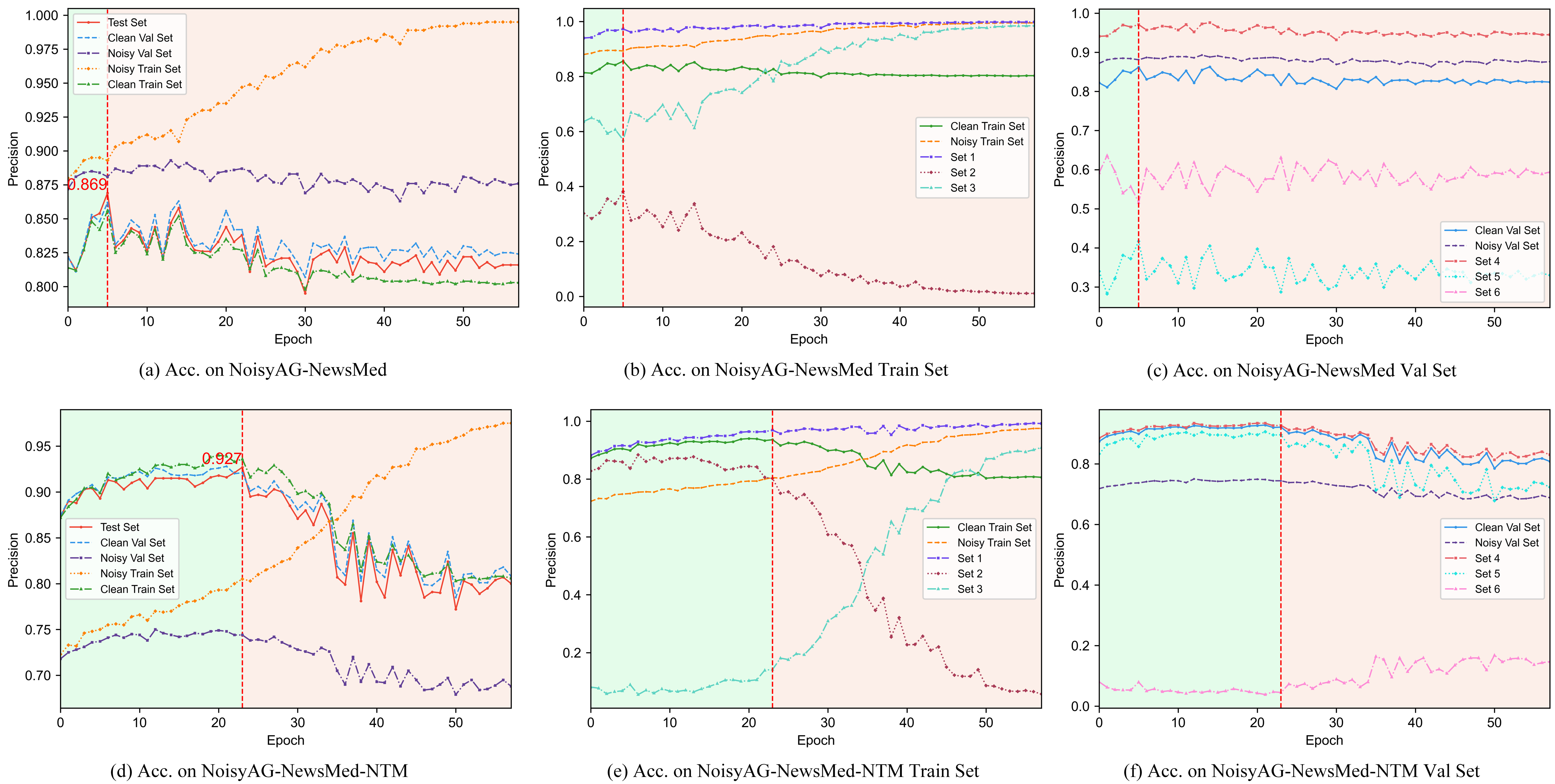}
	\caption{Acc. on NoisyAG-NewsMed and NoisyAG-NewsMed-NTM. }
        \label{fig:medLearn}
\end{figure}

During training, we monitored the classifier's performance across various datasets, including noisy and clean training and validation sets, as well as subsets like Set1 and Set2 illustrated in Figure \ref{fig:dataSplitAndState} (a). Referring to Figure \ref{fig:medLearn}, for synthetic noise, accuracy on Set3 started low and improved, while Set2 started high and decreased, suggesting early fitting of clean samples and later fitting of noisy data, making early stopping beneficial. In contrast, NoisyAG-News showed high initial accuracy for Set3 that stabilized, while Set2's low initial accuracy declined, indicating early fitting of instance-dependent noisy labels.Validation accuracy plots showed that Set3 and Set6 from instance-dependent noise shared the same distribution, unlike synthetic noise. Thus, Set6 accuracy was higher than Set5 for NoisyAG-News, whereas the opposite was true for synthetic noise datasets. This indicates that instance-dependent noise complicates distinguishing confusing samples. The model tends to overfit instance-dependent noise and lacks robustness. Models trained with instance-dependent noise perform well on noisy validation sets due to feature dependence, but their accuracy on clean validation sets is much lower due to conflicting labels between Set6 and Set5. A more detailed analysis can be found in Appendix \ref{supplement:G}.

\subsection{Effect on Different Factor }
\subsubsection{Noise Pattern and Noise Ratio}

Table \ref{tab:PerformanceGap} shows the impact of different noise patterns and rates on model accuracy. Synthetic noise accuracy is comparable to training with clean samples, with accuracy dropping by no more than 3\% even at a 38\% noise ratio. However, noise generated with the same NTM is more challenging, resulting in a 3\% to 7\% accuracy drop. Across all models, the NoisyAG-News dataset shows average accuracy drops of 3.5\%, 8.0\%, and 16.0\% for noise ratios of 10\%, 20\%, and 38\%.

\begin{table*}[]
\centering
\caption{Model Performance Gap Across Different Noise Levels.}
\scalebox{0.72}{
\begin{tabular}{@{}llllllllllllllllll@{}}
\toprule
 & \multicolumn{1}{c}{} & \multicolumn{3}{c}{Ours} & \multicolumn{3}{c}{Flip by NTM} &  \multicolumn{3}{c}{Synth IDN}  &  \multicolumn{3}{c}{Single-Flip}  &  \multicolumn{3}{c}{Uniform} 
 \\ \cmidrule(lr){3-5} \cmidrule(lr){6-8} \cmidrule(lr){9-11}  \cmidrule(lr){12-14} \cmidrule(lr){15-17} 
  & \multicolumn{1}{c}{Clean} & \multicolumn{1}{c}{10\%} & \multicolumn{1}{c}{20\%} & \multicolumn{1}{c}{38\%} & \multicolumn{1}{c}{10\%} & \multicolumn{1}{c}{20\%} & \multicolumn{1}{c}{38\%}  & \multicolumn{1}{c}{10\%} & \multicolumn{1}{c}{20\%} & \multicolumn{1}{c}{38\%} & \multicolumn{1}{c}{10\%} & \multicolumn{1}{c}{20\%} & \multicolumn{1}{c}{38\%} & \multicolumn{1}{c}{10\%} & \multicolumn{1}{c}{20\%} & \multicolumn{1}{c}{38\%} \\ \midrule

BERT & 93.4 & 3.4 & 7.6 & 16.0 & 0.5 & 1.0 & 5.6 & 0.5 & 1.2 & 1.4 & 0.5 & 0.5 & 1.7 & 0.7 & 1.0 & 1.8  \\
RoBERTa & 93.9 & 3.3 & 7.0 & 16.8 & 0.6 & 1.2 & 3.7 & 0.5 & 1.1 & 1.7 & 0.4 & 0.5 & 0.8 & 0.6 & 0.9 & 2.0 \\
DeBERTa-V3 & 93.6 & 3.6 & 7.3 & 15.9 & 0.3 & 0.8 & 3.1 & 0.4 & 0.7 & 1.6 & 0.3 & 0.6 & 0.8 & 0.4 & 0.2 & 1.6 \\
BART & 93.7 & 3.3 & 6.9 & 15.2 & 0.4 & 1.1 & 3.7 & 0.5 & 1.1 & 1.7 & 0.2 & 0.7 & 1.7 & 0.5 & 0.8 & 1.7 \\
XLNET & 93.8 & 3.6 & 7.2 & 16.4 & 1.0 & 1.6 & 7.0 & 1.2 & 1.7 & 2.9 & 0.8 & 1.1 & 2.2 & 1.2 & 0.9 & 2.3 \\
 \bottomrule
\end{tabular}}
\label{tab:PerformanceGap}
\end{table*}

\subsubsection{The Effectiveness of the LNL Method }

We employed several representative noise handling techniques \cite{jindal2019effective,yao2020dual,han2018co,zhang2021delving,wei2021smooth,amid2019robust} in experiments on the NoisyAG-News and synthetic noise datasets. In these experiments, "WN" indicates no noise handling methods were used. As shown in Table \ref{tab:LNL}, in most cases, these methods did not significantly improve performance. As the noise level increased, the accuracy on NoisyAG-News was significantly lower than on the synthetic noise and noise-free datasets. The CMGT method showed a noticeable improvement in accuracy on the NoisyAG-NewsWorst dataset. However, the CT and CM methods performed particularly poorly on the NoisyAG-NewsWorst-NTM dataset. More experimental results can be found in Appendix \ref{supplement:F}.

\begin{table*}[]
\centering
\caption{Comparison of LNL Method Performance Using the RoBERTa Model.}
\scalebox{0.72}{
\begin{tabular}{@{}llllllllllllllllll@{}}
\toprule
 & \multicolumn{1}{c}{} & \multicolumn{3}{c}{Ours} & \multicolumn{3}{c}{Flip by NTM} &  \multicolumn{3}{c}{Synth IDN}  &  \multicolumn{3}{c}{Single-Flip}  &  \multicolumn{3}{c}{Uniform} 
 \\ \cmidrule(lr){3-5} \cmidrule(lr){6-8} \cmidrule(lr){9-11}  \cmidrule(lr){12-14} \cmidrule(lr){15-17} 
  & \multicolumn{1}{c}{Clean} & \multicolumn{1}{c}{10\%} & \multicolumn{1}{c}{20\%} & \multicolumn{1}{c}{38\%} & \multicolumn{1}{c}{10\%} & \multicolumn{1}{c}{20\%} & \multicolumn{1}{c}{38\%}  & \multicolumn{1}{c}{10\%} & \multicolumn{1}{c}{20\%} & \multicolumn{1}{c}{38\%} & \multicolumn{1}{c}{10\%} & \multicolumn{1}{c}{20\%} & \multicolumn{1}{c}{38\%} & \multicolumn{1}{c}{10\%} & \multicolumn{1}{c}{20\%} & \multicolumn{1}{c}{38\%} \\ \midrule

WN & 93.4 & 90.0 & 85.8 & 77.4 & 92.9 & 92.4 & 87.8 & 92.9 & 92.2 & 92.0 & 92.9 & 92.9 & 91.7 & 92.7 & 92.4 & 91.6  \\
CM \cite{jindal2019effective} & 92.4 & 90.1 & 86.7 & 79.5 & 92.8  & 90.3 & 70.0 & 91.7 & 90.7 & 90.2 & 92.4 & 91.7 & 90.2  & 91.8 & 91.2 & 90.8\\ 
CMGT \cite{yao2020dual} & 93.9 & 90.4 & 87.5 & 84.6 & 93.5 & 93.2 & 90.7 & 93.5 & 92.9 & 92.3 & 93.5 & 93.2 & 93.3 & 93.4 & 93.0 & 92.1\\
CT \cite{han2018co} & 94.0 & 90.4 & 86.9 & 77.1 & 93.1 & 92.4 & 77.9 & 93.2 & 93.1 & 91.8  & 93.4 & 93.0 & 91.6 & 93.4 &93.2 & 91.6\\
LS  \cite{zhang2021delving}& 93.9 & 90.5 & 86.8 & 77.0 & 93.4 & 92.7 & 90.2 & 93.4 & 92.8 & 91.8 & 93.6 & 93.0 & 92.6 & 93.4 & 93.2 & 91.7\\
NLS \cite{wei2021smooth} & 93.9 & 90.5 & 86.8 & 77.0 & 93.3 & 92.6  & 90.5 & 93.1 & 92.9 & 91.9 & 93.5 & 93.0 & 92.5 & 93.2 & 93.2 & 92.1\\
BTLS \cite{amid2019robust} & 93.8 & 89.7 & 86.2 & 77.3 & 93.3 & 92.6 & 90.6 & 93.2 & 92.7 & 91.6 & 93.2 & 93.0 & 92.1 & 93.2 & 93.1 & 91.9\\
 \bottomrule
\end{tabular}}
\label{tab:LNL}
\end{table*}

\section{Conclusion}

Based on the AG-News dataset, we present NoisyAG-News, an easy-to-use benchmark for noisy text classification tasks. We qualitatively and quantitatively demonstrate that human noise is feature-dependent and significantly different from synthetic label noise. Experiments on NoisyAG-News and corresponding synthetic datasets, using various models and noise handling methods, revealed new findings:

\begin{itemize}
	\item[$\bullet$] Human noise shows more complex patterns and is harder to handle than synthetic noise.
	\item[$\bullet$] For synthetic noise, models fit true-labeled samples early, making early stopping effective. In contrast, for instance-dependent noise in NoisyAG-News, models struggle to distinguish between good and bad samples, reducing prediction accuracy.
        \item[$\bullet$] In text classification with pre-trained models, most noise handling methods are less effective for both synthetic and instance-dependent noise. Methods incorporating prior information are most effective.
        
\end{itemize}

This study provides valuable insights and a solid benchmark for advancing research in noisy text classification and robust learning from noisy labels.

\bibliography{references}

\newpage
\appendix

\section{Analysis of Annotated Labels}
\label{supplement:A}
% 标注过程和三个标注的质量分析

We analyzed the three manually annotated labels to demonstrate how different individuals perceive the same sample, revealing the instance-dependency of both annotated and synthetic labels.

Initially, all samples were copied three times and then distributed to three groups for annotation, with each group consisting of 20 annotators. Thus, for each sample, three labels were obtained. To control annotation quality, we calculated the accuracy of each annotator and the consistency of labels across different groups. Once the quality requirements were met, the sample labels were inferred from the three manual annotations using different methods, resulting in datasets with varying noise rates.

Prior to large-scale annotation, a small-scale annotation of 4,000 samples was conducted to ensure the process was sound. After confirming the procedure, a large-scale annotation of 46,000 samples was carried out. The accuracy and consistency of the annotations are shown in the following table \ref{tab:sampleAcc}.

\begin{table}[htbp]
\centering
\caption{Accuracy on Different Sample Set.}
\label{tab:sampleAcc}
% 增加行间距，调整为1.5或根据需要选择合适的数值
\renewcommand{\arraystretch}{1.5}
\begin{tabular}{ccccc}
\toprule
& group 1 & group 2 & group 3  \\
\toprule
sample 4000  & 78.1 & 78.0 & 76.8   \\
\midrule
sample 50000  & 77.2  & 76.6 & 78.7  \\
\bottomrule
\end{tabular}
\end{table}

\begin{table}[htbp]
\centering
\caption{The Consistency of Annotations across Different Groups.}
\label{tab:ConsistencyGroup}
% 增加行间距，调整为1.5或根据需要选择合适的数值
\renewcommand{\arraystretch}{1.5}
\begin{tabular}{ccccc}
\toprule
& Observed Score & Cohen's Kappa & Gwet's Gamma  \\
\toprule
sample 4000  & 0.81 & 0.75 & 0.75   \\
\midrule
sample 50000  & 0.79  & 0.72 & 0.72  \\
\bottomrule
\end{tabular}
\end{table}

The accuracy for each group exceeded 75\%, and the consistency surpassed 0.7 in Table \ref{tab:ConsistencyGroup}, meeting the annotation requirements. We observed that as the number of samples requiring annotation increased, both accuracy and consistency showed varying degrees of decline. This suggests that annotator concentration may affect judgment.

\begin{figure}
	\centering
	\includegraphics[width=14cm]{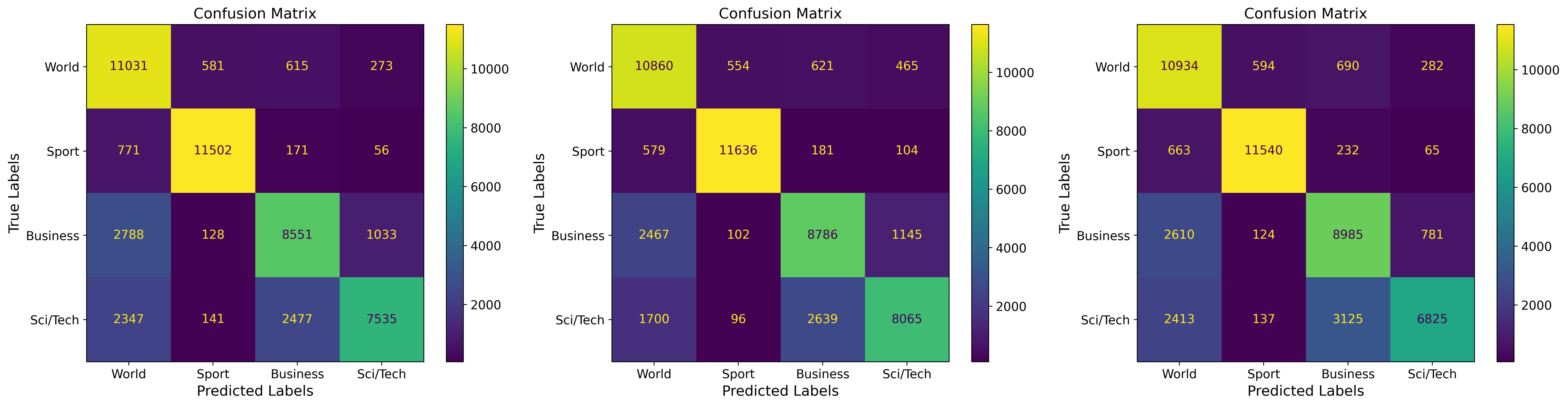}
	\caption{ Confusion of Different Group.  }
        \label{fig:groupConfusion}
\end{figure}

Next, we present the confusion matrices for annotations from different groups in Fig \ref{fig:groupConfusion}. It can be observed that they follow the same pattern: the annotation accuracy for the categories Sport and World is significantly higher than for Business and Sci/Tech. A substantial number of samples with a GT label of Business were annotated as World, and samples with a GT label of Sci/Tech were annotated as World or Business. This indicates that the annotation accuracy varies across different categories and that errors are not uniformly distributed. Instead, they tend to flip to categories that are more related to the current confusing instance.

\begin{figure}
	\centering
	\includegraphics[width=14cm]{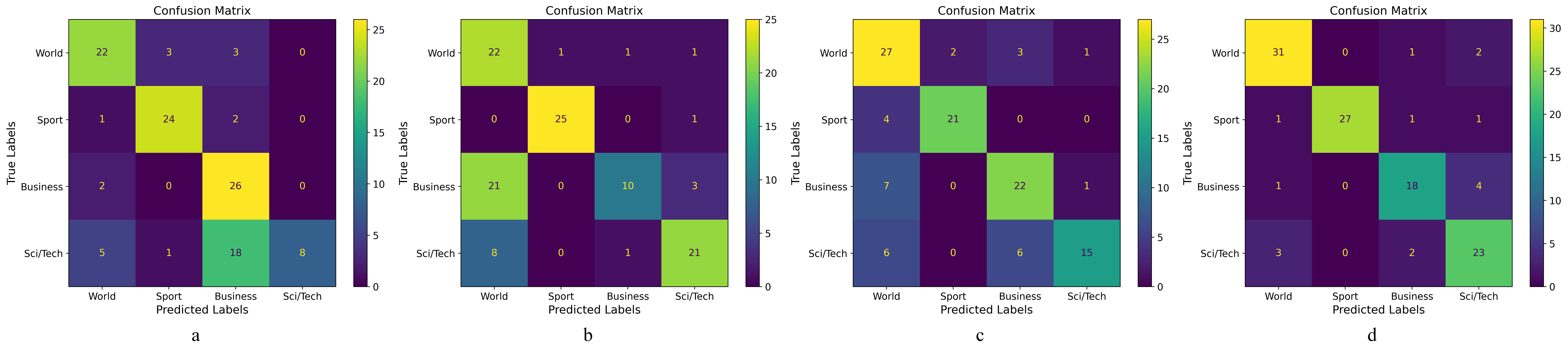}
	\caption{ Confusion of Different Annotator.  }
        \label{fig:AnnotatorConfusion}
\end{figure}

We sampled 5\% of the annotated samples from each annotator and calculated the confusion matrices to observe the characteristics of different annotators. As shown in Figure \ref{fig:AnnotatorConfusion}, the four annotators performed consistently for the World and Sport. However, for the Business, annotator A had a recall of 92.8\% and a precision of 53.0\%, while annotator B had a recall of 29.4\% and a precision of 83.3\%. For the Sci/Tech class, annotator A had a recall of 25\% and a precision of 100\%, while annotator B had a recall of 70\% and a precision of 80.7\%. The performances of annotators A and B were markedly different. Annotator A tended to label confusing samples between Business and Sci/Tech as Business, whereas annotator B tended to label these confusing samples as World. Annotators C and D generally aligned with the overall annotation group, but annotator D was more accurate in judging Sci/Tech samples. For confusing samples in the Business class, annotator C tended to label them as World, while annotator D tended to label them as Sci/Tech.

By observing the performance of these four annotators, we can conclude that each annotator's cognitive and judgment levels are inconsistent, especially for confusing samples. Annotators label samples based on the characteristics of these confusing instances, further indicating that the noise labels generated from annotation are instance-dependent.

% \textcolor{red}{why noise high }

The confusion matrices indicate that the precision and recall for each class vary, reflecting the inconsistent levels of confusion among different classes. We observed the confusion degree of samples in different classes by plotting the density curves of consistency scores. For each sample with three labels, if all three labels are identical, the consistency score is 1. If two labels are the same, the consistency score is 0.5. If all three labels differ, the consistency score is 0. The density curves based on these consistency scores are shown below.

\begin{figure}
	\centering
	\includegraphics[width=10cm]{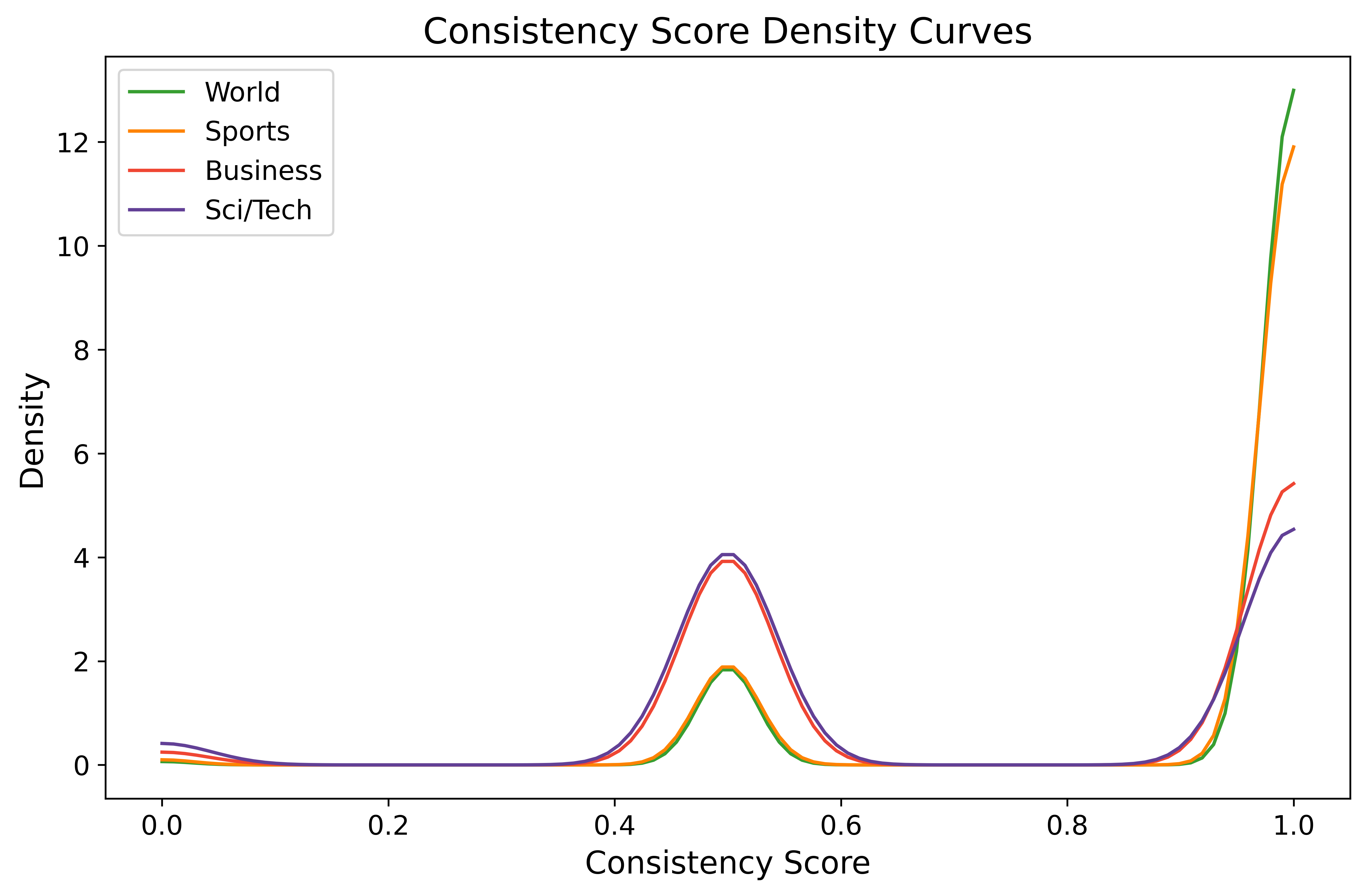}
	\caption{ Consistency Score Density Curves for Different Class.  }
        \label{fig:DensityCurves}
\end{figure}

The density curves indicate \ref{fig:DensityCurves} that there are very few samples with completely inconsistent labels, suggesting that annotators generally took their task seriously. For the Business and Sci/Tech classes, the number of samples with consistency scores of 0.5 and 1 are relatively close. In contrast, the World and Sport classes exhibit higher consistency, with most samples having identical labels. This indicates that the level of confusion among samples varies across different classes, leading to inconsistent annotations and instance-dependent noise.

\section{ NTMs of NoisyAG-News and Corresponding SyntheticNoise}
\label{supplement:B}

\begin{figure}
	\centering
	\includegraphics[width=14cm]{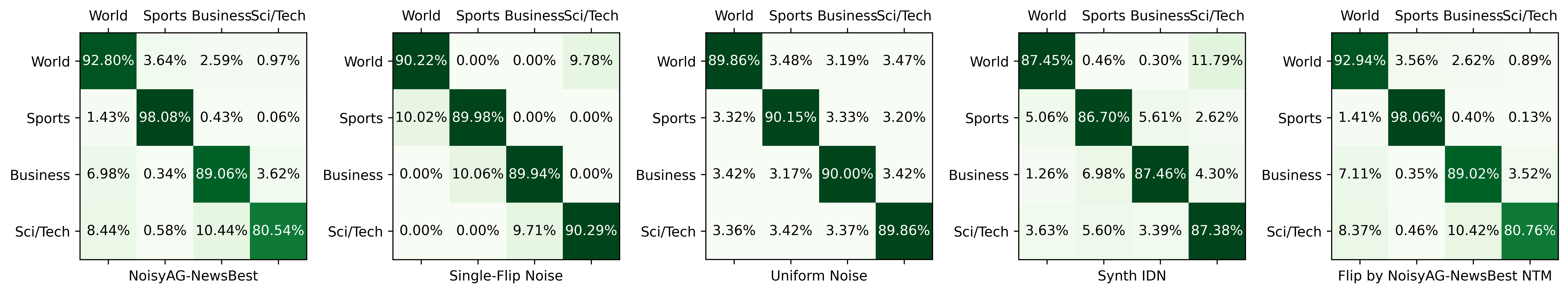}
	\caption{ NTMs for NoisyAg-NewsBest and Corresponding SyntheticNoise.  }
        \label{fig:BestNTM}
        
\end{figure}

\begin{figure}
	\centering
	\includegraphics[width=14cm]{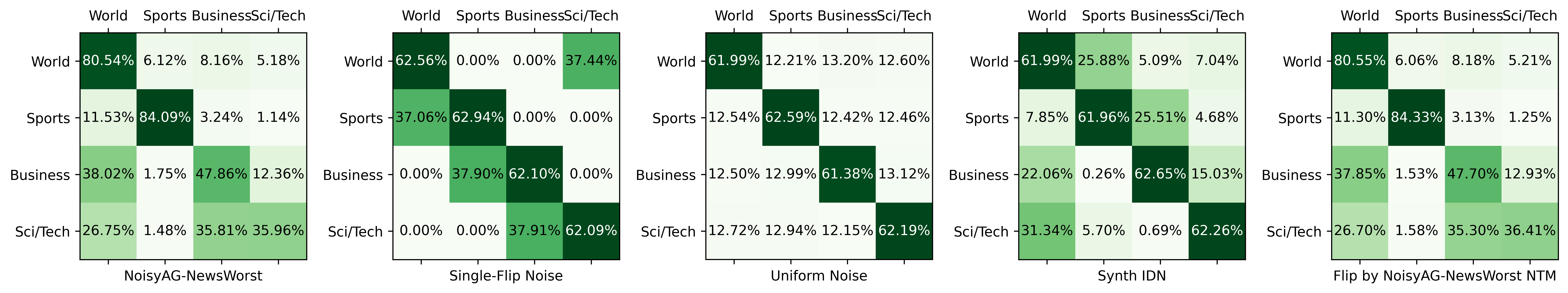}
	\caption{ NTMs for NoisyAg-NewsWorst and Corresponding SyntheticNoise.   }
        \label{fig:WorstNTM}
\end{figure}

We present a comparison of the noise transition matrices for NoisyAG-NewsBest \ref{fig:BestNTM}, NoisyAG-NewsWorst \ref{fig:WorstNTM}, and their corresponding synthetic noises.
% 主要还是画两张图。

\section{ Qualitative and Quantitative Evidence}
\label{supplement:D}

\subsection{Qualitative Evidence}

\begin{figure}
	\centering
	\includegraphics[width=12cm]{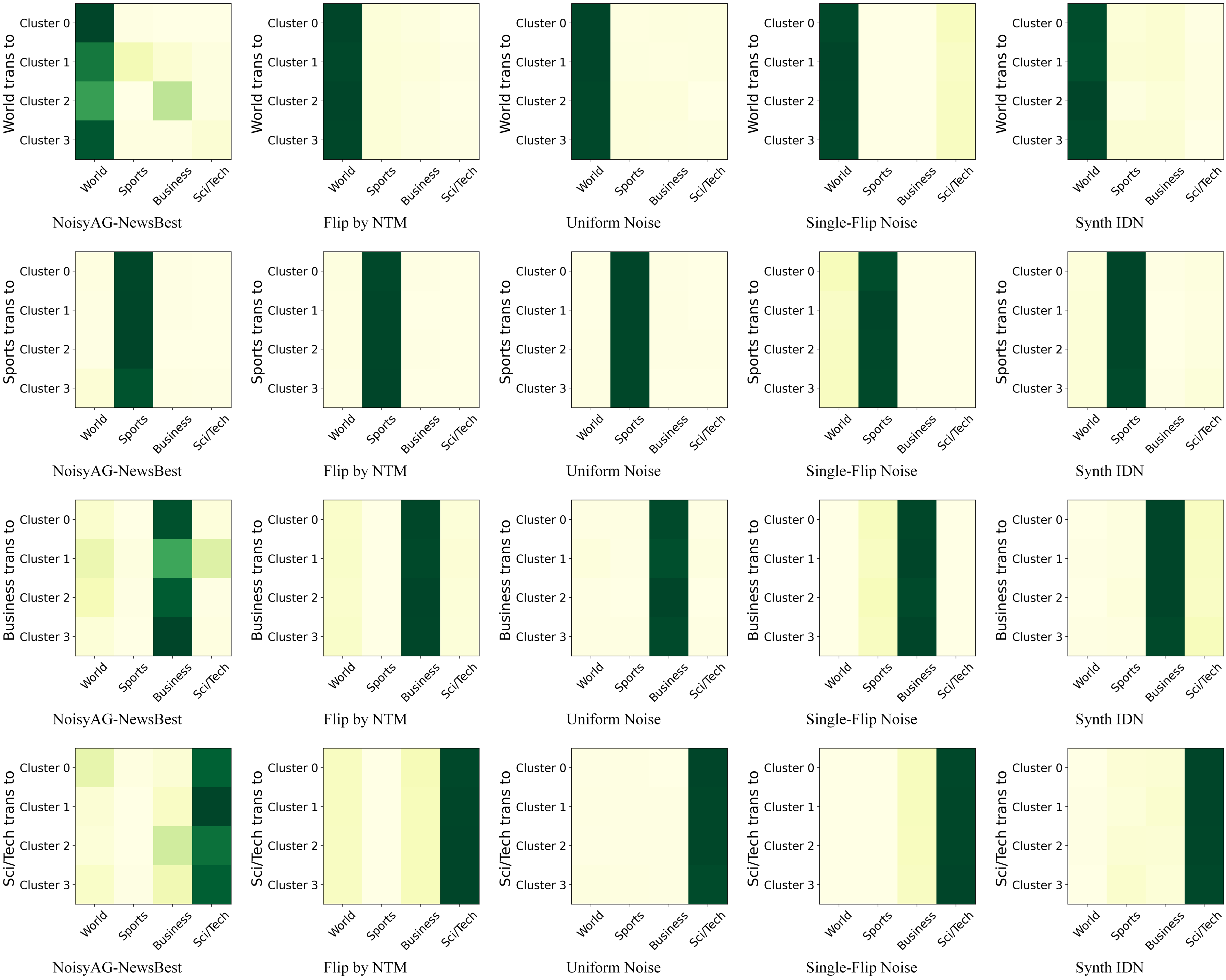}
	\caption{  \( {p}_{i, \nu}[j]\) in NoisyAG-NewsBest and Synthetic Noise.  }
        \label{fig:BestPV}
\end{figure}

\begin{figure}
	\centering
	\includegraphics[width=12cm]{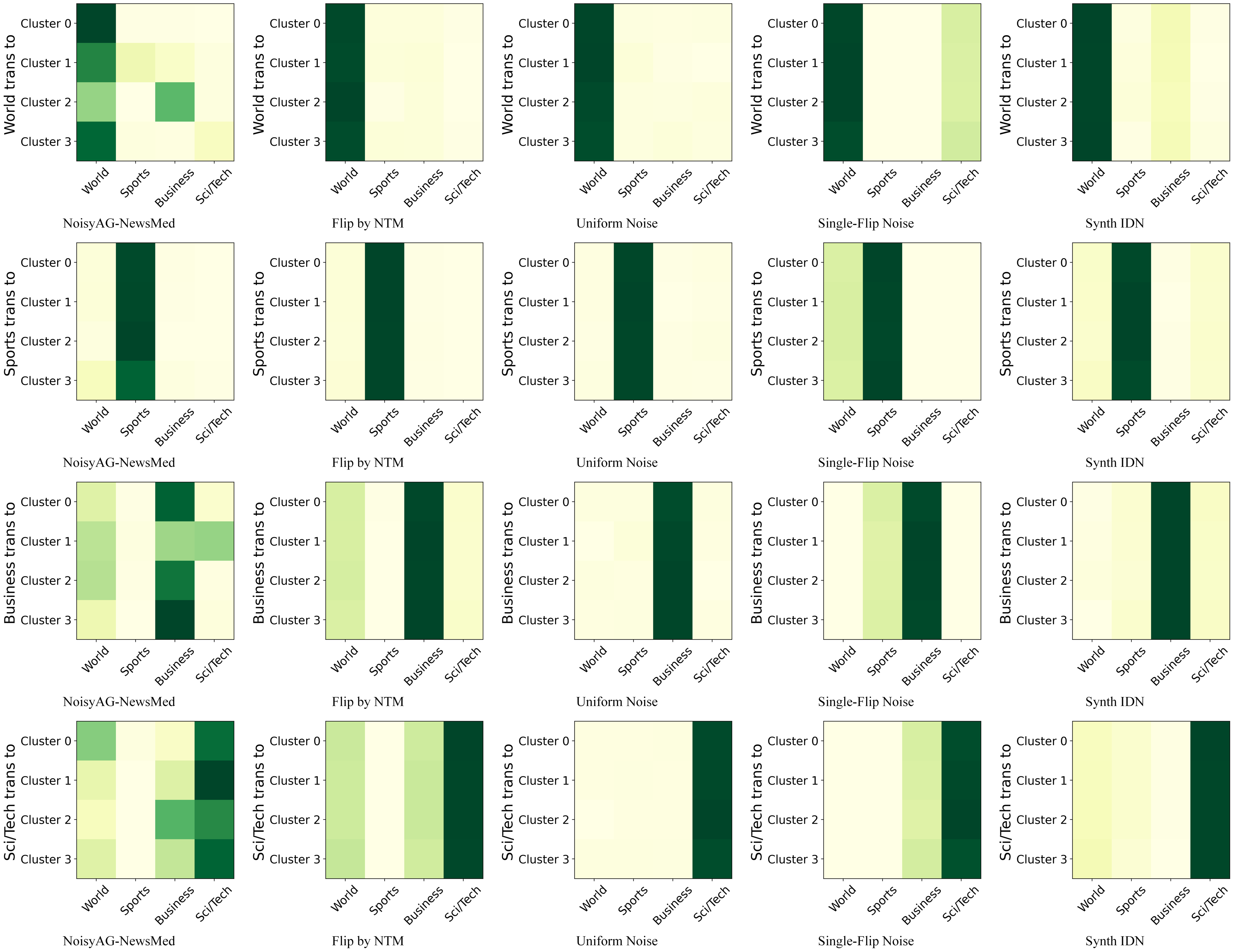}
	\caption{ \( {p}_{i, \nu}[j]\) in NoisyAG-NewsMed and Synthetic Noise.  }
        \label{fig:MedPV}
\end{figure}

\begin{figure}
	\centering
	\includegraphics[width=12cm]{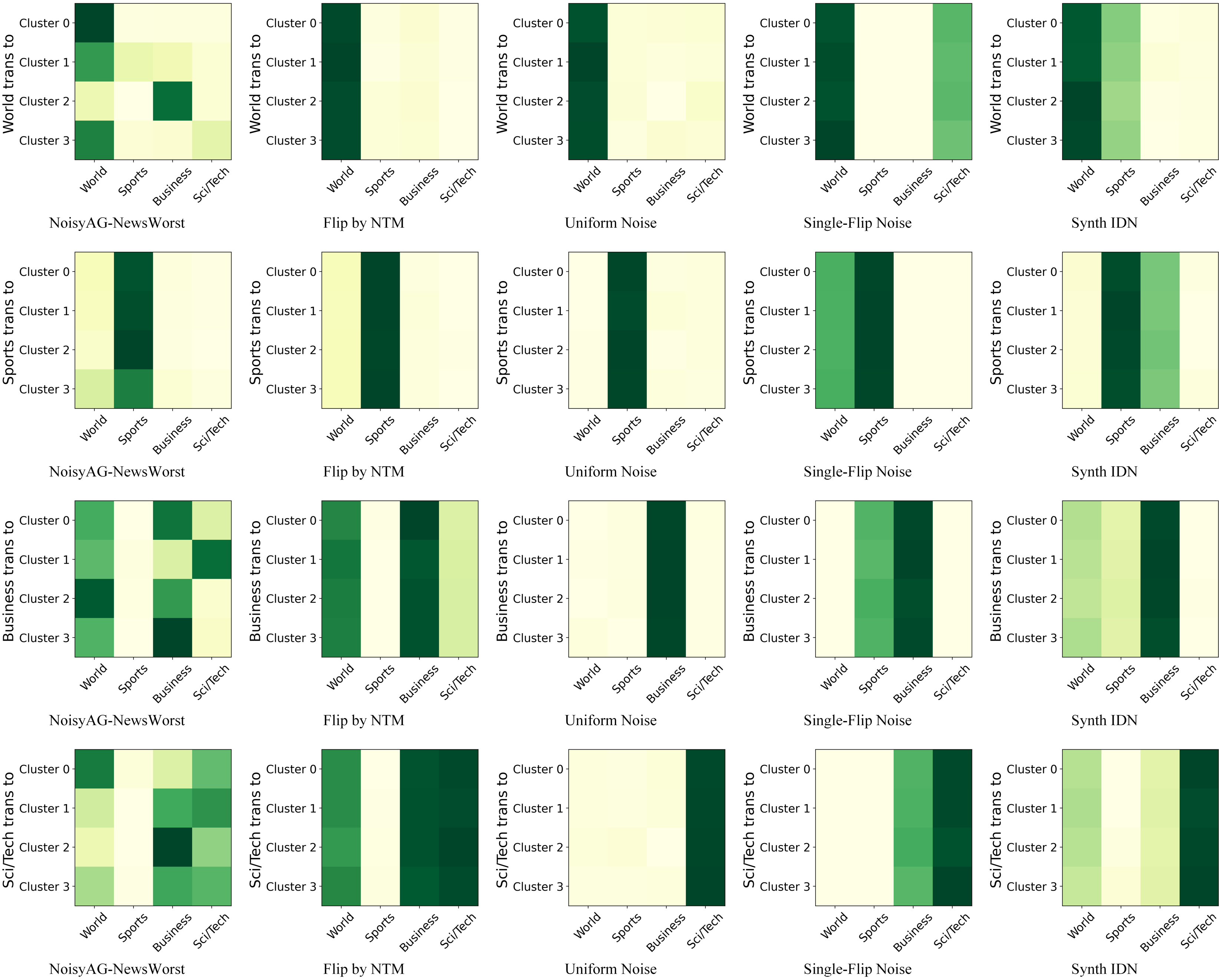}
	\caption{ \( {p}_{i, \nu}[j]\) in NoisyAG-NewsWorst and Synthetic Noise.  }
	\label{fig:WorstPV}
\end{figure}

We present the transition vectors  \( {p}_{i, \nu}[j]\) for each class in NoisyAG-News and its corresponding synthetic matrices \ref{fig:BestPV} \ref{fig:MedPV} \ref{fig:WorstPV}. The figures show that for each dataset in NoisyAG-News, the transition vectors between different clusters of each class vary significantly. In contrast, the differences are minimal for synthetic noise, demonstrating that NoisyAG-News is feature-dependent.

\subsection{Quantitative Evidence}

\begin{table}[htbp]
\small 
\centering
\caption{ \( h, s \), and \( p\)-value between NoisyAG-NewsBest and Different Synthetic Noise.   }
\label{tab:BestQuantitative}
\scalebox{0.76}{
% 增加行间距，调整为1.5或根据需要选择合适的数值
\renewcommand{\arraystretch}{1.2}
\begin{tabular}{ccccc}
\toprule
\multirow{2}{*}{Class} & \multicolumn{4}{c}{Noise type} \\
\cline{2-5}
& Flip by NTM & Uniform & Single-Flip & Synth-IDN \\
\toprule
Total  & 1.0373 / 8.2251/ \(2.1\times 10^{-34}\) & 0.8828 / 11.2948/ \( 6.8\time 10^{-36} \) & 0.7467 / 13.3264 /\( 4.39\times 10^{-37}\) & 4.8645 / 12.435 / \(  7.1\time 10^{-19} \)  \\
\toprule
World  & 1.2161 / 14.0148  & 1.011 / 13.3803 & 1.1102 / 15.8159 & 5.2674 / 14.6708  \\
\midrule
Sports  & 0.3778 / 1.4208  & 0.7869 / 6.9219  & 0.6633 / 8.8794 & 3.9386 / 10.2978   \\
\midrule
Business  & 1.1884 / 9.9467 & 0.857 / 11.6114  & 0.6208 / 15.963 & 6.2909 / 12.3682 \\
\midrule
Sci/Tech  & 1.3667 / 7.518 & 0.8763 / 13.2654 & 0.5926 / 12.6473 & 3.9611 / 12.4032 \\
\bottomrule
\end{tabular}}
\end{table}

\begin{table}[htbp]
\small 
\centering
\caption{ \( h, s \), and \( p\)-value between NoisyAG-NewsMed and Different Synthetic Noise.   }
\label{tab:MedQuantitative}
\scalebox{0.76}{
% 增加行间距，调整为1.5或根据需要选择合适的数值
\renewcommand{\arraystretch}{1.2}
\begin{tabular}{ccccc}  
\toprule
\multirow{2}{*}{Class} & \multicolumn{4}{c}{Noise type} \\
\cline{2-5}
& Flip by NTM & Uniform & Single-Flip & Synth-IDN \\
\toprule
Total  & 1.1591 / 13.0671/ \(4.3\times 10^{-36}\)  & 1.3631 / 20.1451/\(  6.3\time 10^{-40} \) & 0.9532 / 24.3418/ \( 4.13\times 10^{-39}\) & 7.7526 / 21.444 /\( 5.0\times 10^{-17}\)   \\
\toprule
World  & 1.376 / 21.8387 & 2.0222 / 20.4328 & 1.1894 / 26.1243 & 8.9577 / 22.0014  \\
\midrule
Sports  & 0.5215 / 3.1585  & 1.0337 / 13.3219  & 0.7651 / 16.0614 & 8.2038 / 15.7378   \\
\midrule
Business  & 1.2158 / 15.3427 & 1.1403 / 22.3826  & 0.8768 / 31.3167 & 6.9934 / 22.8905 \\
\midrule
Sci/Tech  & 1.523 / 11.9286 & 1.2562 / 24.443 & 0.9815 / 23.8647 & 6.8556 / 25.1462 \\
\bottomrule
\end{tabular}}
\end{table}

\begin{table}[htbp]
\small 
\centering
\caption{ \( h, s \), and \( p\)-value between NoisyAG-NewsWorst and Different Synthetic Noise.   }
\label{tab:WorstQuantitative}
\scalebox{0.76}{
% 增加行间距，调整为1.5或根据需要选择合适的数值
\renewcommand{\arraystretch}{1.2}
\begin{tabular}{ccccc}
\toprule
\multirow{2}{*}{Class} & \multicolumn{4}{c}{Noise type} \\
\cline{2-5}
& Flip by NTM & Uniform & Single-Flip & Synth-IDN \\
\toprule
Total  & 1.7201 / 16.7765/ \(2.2\times 10^{-36}\) & 1.7338 / 29.4002/ \( 4.8\time 10^{-41} \) &1.184 / 37.5088/ \( 1.3\times 10^{-44}\) & 18.4851 / 32.514/\(4.1\times 10^{-11}\) \\
\toprule
World  & 1.978 / 29.1228 & 2.2636 / 26.4742 & 1.2714 / 39.8231 & 22.7464 / 30.4509  \\
\midrule
Sports  & 1.2696 / 5.487  & 1.3855 / 20.5945  & 1.2685 / 25.5087 & 20.4842 / 25.6267   \\
\midrule
Business  & 1.6946 / 16.4254 & 1.7874 / 33.5546  & 1.2784 / 51.3678 & 15.446 / 35.1867 \\
\midrule
Sci/Tech  & 1.9381 / 16.0707 & 1.4988 / 36.9777 & 0.9178 / 33.3355 & 15.2638 / 38.7919 \\
\bottomrule
\end{tabular}}
\end{table}

As a supplement to Table \ref{tab:Quantitative}, we present the average distance statistics \( h \) and \( s \) for each class in the NoisyAG-News dataset and its corresponding synthetic noise datasets \ref{tab:BestQuantitative} \ref{tab:MedQuantitative} \ref{tab:WorstQuantitative}. We obtained an extremely small p-value (p < 0.0001), demonstrating that our dataset is different from synthetic noise and is instance-dependent. It can be observed that, compared to other synthetic noises, the distance \( s \) of Synth-IDN is closer to our distance \( h \). For the same dataset, such as NoisyAG-NewsMed, the \( h \) and \( s \) distributions for the Sports category often exhibit less variance than other categories, likely because Sports samples are easier to distinguish. Comparing the NoisyAG-News datasets horizontally, we find that as the noise rate increases, the difference between the instance-dependent noise distance \( h \) and the synthetic noise distance \( s \) becomes larger, indicating that the noise in the dataset becomes more instance-dependent with higher noise rates.

\section{Detailed Experimental Setup}
\label{supplement:E}

We conducted the training on a server equipped with eight A6000 Ada GPUs, setting the batch size to 32 and using the AdamW optimizer to minimize the loss. The initial learning rate was set to 2e-5 and decayed progressively during training. We set the maximum training steps to 20,000 and implemented early stopping. All these settings will be reflected in our open-source code.

\section{More Result}
\label{supplement:F}

We presented the performance of BERT \ref{tab:BERTAcc}, XLNET \ref{tab:XLNETAcc}, DeBERTa \ref{tab:DeBERTaAcc} , and BART \ref{tab:BARTAcc} models using different noise-handling methods to address the noise in NoisyAG-News and its corresponding synthetic noise.

\begin{table*}[]
	\centering
	\caption{Comparison of LNL Method Performance Using the BERT Model.}
	\label{tab:BERTAcc}
	\scalebox{0.72}{
		\begin{tabular}{@{}llllllllllllllllll@{}}
			\toprule
			& \multicolumn{1}{c}{} & \multicolumn{3}{c}{Ours} & \multicolumn{3}{c}{Flip by NTM} &  \multicolumn{3}{c}{Synth IDN}  &  \multicolumn{3}{c}{Single-Flip}  &  \multicolumn{3}{c}{Uniform} 
			\\ \cmidrule(lr){3-5} \cmidrule(lr){6-8} \cmidrule(lr){9-11}  \cmidrule(lr){12-14} \cmidrule(lr){15-17} 
			& \multicolumn{1}{c}{Clean} & \multicolumn{1}{c}{10\%} & \multicolumn{1}{c}{20\%} & \multicolumn{1}{c}{38\%} & \multicolumn{1}{c}{10\%} & \multicolumn{1}{c}{20\%} & \multicolumn{1}{c}{38\%}  & \multicolumn{1}{c}{10\%} & \multicolumn{1}{c}{20\%} & \multicolumn{1}{c}{38\%} & \multicolumn{1}{c}{10\%} & \multicolumn{1}{c}{20\%} & \multicolumn{1}{c}{38\%} & \multicolumn{1}{c}{10\%} & \multicolumn{1}{c}{20\%} & \multicolumn{1}{c}{38\%} \\ \midrule 
			
			WN & 93.4 & 90.0 & 86.1 & 77.7 & 92.9 & 92.4 & 87.9 & 92.8 & 92.2 & 91.0 & 92.8 & 92.8 & 91.7 & 92.7 & 92.4 & 91.6  \\
			CM & 92.6 & 89.8 & 86.0 & 80.1 & 92.4 & 90.9 & 70.1 & 92.3 & 91.7 & 90.9 & 92.2 & 91.8 & 90.1 & 92.7 & 91.7 & 91.2 \\ 
			CMGT & 93.3 & 90.3 & 87.1 & 83.0 & 93.0 & 92.4 & 90.8 & 92.7 & 92.5 & 91.8 & 93.0 & 92.6 & 92.8 & 92.8 & 92.8 & 91.5\\
			CT & 93.5 & 90.1 & 86.2 & 77.1 & 92.7 & 91.6 & 75.2 & 92.8 & 92.4 & 91.2 & 92.8 & 92.6 & 91.0 & 93.0 & 92.4 & 91.6        \\
			LS & 93.4 & 90.0 & 86.3 & 76.6 & 93.0 & 92.2 & 87.0 & 92.8 & 92.7 & 91.5 & 92.8 & 92.5 & 91.5 & 93.0 & 92.8 & 91.4 \\
			NLS & 93.4 & 89.9 & 86.2 & 77.5 & 93.0 & 92.3 & 89.0 & 92.7 & 92.6 & 91.8 & 92.8 & 92.4 & 91.5 & 92.9 & 92.6 & 91.8 \\
			BTLS &93.1 & 90.2 & 86.3 & 76.5 & 92.7 & 92.1 & 87.5 & 92.8 & 92.4 & 91.1 & 92.8 & 92.2 & 91.0 & 92.5 & 92.3 & 91.4\\
			\bottomrule
	\end{tabular}}

\end{table*}

\begin{table*}[]
	\centering
	\caption{Comparison of LNL Method Performance Using the XLNET Model.}
	\label{tab:XLNETAcc}
	\scalebox{0.72}{
		\begin{tabular}{@{}llllllllllllllllll@{}}
			\toprule
			& \multicolumn{1}{c}{} & \multicolumn{3}{c}{Ours} & \multicolumn{3}{c}{Flip by NTM} &  \multicolumn{3}{c}{Synth IDN}  &  \multicolumn{3}{c}{Single-Flip}  &  \multicolumn{3}{c}{Uniform} 
			\\ \cmidrule(lr){3-5} \cmidrule(lr){6-8} \cmidrule(lr){9-11}  \cmidrule(lr){12-14} \cmidrule(lr){15-17} 
			& \multicolumn{1}{c}{Clean} & \multicolumn{1}{c}{10\%} & \multicolumn{1}{c}{20\%} & \multicolumn{1}{c}{38\%} & \multicolumn{1}{c}{10\%} & \multicolumn{1}{c}{20\%} & \multicolumn{1}{c}{38\%}  & \multicolumn{1}{c}{10\%} & \multicolumn{1}{c}{20\%} & \multicolumn{1}{c}{38\%} & \multicolumn{1}{c}{10\%} & \multicolumn{1}{c}{20\%} & \multicolumn{1}{c}{38\%} & \multicolumn{1}{c}{10\%} & \multicolumn{1}{c}{20\%} & \multicolumn{1}{c}{38\%} \\ \midrule 
			
			WN & 93.8 & 90.2 & 86.6 & 77.4 & 93.4 & 92.7 & 90.3 & 93.1 & 92.7 & 91.3 & 93.2 & 93.2 & 92.3 & 93.0 & 92.8 & 92.1 \\
			CM & 92.4 & 89.6 & 87.2 & 83.6 & 92.5 & 90.7 & 66.3 & 92.4 & 91.9 & 90.4 & 92.3 & 91.5 & 90.1 & 92.0 & 90.9 & 90.8 \\ 
			CMGT & 93.5 & 90.4 & 87.9 & 83.0 & 93.1 & 92.7 & 90.8 & 93.2 & 93.3 & 92.0 & 93.2 & 93.3 & 93.0 & 93.2 & 92.8 & 91.8\\
			CT  & 93.6 & 90.2 & 86.5 & 77.4 & 93.0 & 92.1 & 74.1 &  93.2 & 92.4 & 91.6 & 93.5 & 93.1 & 91.0  & 93.5 & 92.9 & 91.6 \\
			LS  & 93.6 & 90.3 & 85.8 & 77.1 & 93.3 & 92.7 & 90.3 & 93.1 & 93.0 & 91.7 & 93.2 & 93.0 & 92.1 & 93.3 & 93.1 & 92.0 \\
			NLS & 93.8 & 90.1 & 87.5 & 77.7 & 93.2 & 92.6 & 89.5 & 93.1 & 93.0 & 91.9 & 93.4 & 93.3 & 92.6 & 93.1 & 92.9 & 92.0 \\
			BTLS & 93.6 & 90.4 & 86.4 & 77.0 & 92.9 & 92.2 & 88.8 & 93.2 & 92.7 & 91.8 & 93.3 & 93.1 & 91.9 & 93.4 & 93.0 & 91.8\\
			\bottomrule
	\end{tabular}}
	
\end{table*}

\begin{table*}[]
	\centering
	\caption{Comparison of LNL Method Performance Using the DeBERTa-V3 Model.}
	\label{tab:DeBERTaAcc}
	\scalebox{0.72}{
		\begin{tabular}{@{}llllllllllllllllll@{}}
			\toprule
			& \multicolumn{1}{c}{} & \multicolumn{3}{c}{Ours} & \multicolumn{3}{c}{Flip by NTM} &  \multicolumn{3}{c}{Synth IDN}  &  \multicolumn{3}{c}{Single-Flip}  &  \multicolumn{3}{c}{Uniform} 
			\\ \cmidrule(lr){3-5} \cmidrule(lr){6-8} \cmidrule(lr){9-11}  \cmidrule(lr){12-14} \cmidrule(lr){15-17} 
			& \multicolumn{1}{c}{Clean} & \multicolumn{1}{c}{10\%} & \multicolumn{1}{c}{20\%} & \multicolumn{1}{c}{38\%} & \multicolumn{1}{c}{10\%} & \multicolumn{1}{c}{20\%} & \multicolumn{1}{c}{38\%}  & \multicolumn{1}{c}{10\%} & \multicolumn{1}{c}{20\%} & \multicolumn{1}{c}{38\%} & \multicolumn{1}{c}{10\%} & \multicolumn{1}{c}{20\%} & \multicolumn{1}{c}{38\%} & \multicolumn{1}{c}{10\%} & \multicolumn{1}{c}{20\%} & \multicolumn{1}{c}{38\%} \\ \midrule 
			
			WN & 93.6 & 90.3 & 86.3 & 77.7 & 93.3 & 92.8 & 90.5 & 93.2 & 92.9 & 92.0 & 93.3 & 93.0 & 92.8 & 93.2 & 93.4 & 92.0 \\ 
			CM & 91.3 & 89.2 & 87.4 & 81.1 & 92.5 & 90.3 & 48.9 & 91.0 & 91.0 & 90.4 & 91.4 & 91.5 & 89.5 & 90.9 & 91.0 & 90.8 \\ 
			CMGT & 93.8 & 90.0 & 87.0 & 82.5 & 93.1 & 93.0 & 90.7 &  93.2 & 93.3 & 92.4 & 93.2 & 93.3 & 93.2 & 93.1 & 93.0 & 91.9 \\
			CT  & 93.6 & 90.2 & 87.0 & 77.6 & 93.0 & 91.6 & 74.6 & 92.9 & 93.0 & 92.0 & 93.3 & 92.4 & 91.6 & 93.1 & 92.8 & 92.9  \\
			LS  & 93.5 & 90.2 & 86.3 & 77.0 & 93.4 & 92.7 & 90.4 & 93.2 & 92.8 & 92.0 & 93.4 & 92.2 & 91.8 & 93.3 & 93.0 & 92.4 \\
			NLS & 93.5 & 90.1 & 86.7 & 77.2 & 93.2 & 92.6 & 90.1 & 93.2 & 92.6 & 91.9 & 93.2 & 92.8 & 91.8 & 93.2 & 92.4 & 91.6 \\
			BTLS & 93.6 & 90.1 & 86.3 & 78.9 & 93.0 & 91.7 & 90.2 & 93.0 & 93.0 & 92.8 & 93.3 & 93.0 & 92.6 & 93.1 & 92.6 & 91.9 \\
			\bottomrule
	\end{tabular}}

\end{table*}

\begin{table*}[]
	\centering
	\caption{Comparison of LNL Method Performance Using the BART Model.}
	\label{tab:BARTAcc}
	
	\scalebox{0.72}{
		\begin{tabular}{@{}llllllllllllllllll@{}}
			\toprule
			& \multicolumn{1}{c}{} & \multicolumn{3}{c}{ours} & \multicolumn{3}{c}{Flip by NTM} &  \multicolumn{3}{c}{Synth IDN}  &  \multicolumn{3}{c}{Single-Flip}  &  \multicolumn{3}{c}{Uniform} 
			\\ \cmidrule(lr){3-5} \cmidrule(lr){6-8} \cmidrule(lr){9-11}  \cmidrule(lr){12-14} \cmidrule(lr){15-17} 
			& \multicolumn{1}{c}{clean} & \multicolumn{1}{c}{10\%} & \multicolumn{1}{c}{20\%} & \multicolumn{1}{c}{38\%} & \multicolumn{1}{c}{10\%} & \multicolumn{1}{c}{20\%} & \multicolumn{1}{c}{38\%}  & \multicolumn{1}{c}{10\%} & \multicolumn{1}{c}{20\%} & \multicolumn{1}{c}{38\%} & \multicolumn{1}{c}{10\%} & \multicolumn{1}{c}{20\%} & \multicolumn{1}{c}{38\%} & \multicolumn{1}{c}{10\%} & \multicolumn{1}{c}{20\%} & \multicolumn{1}{c}{38\%} \\ \midrule 
			
			WN & 93.7 & 90.4 & 86.8 & 78.5 & 93.3 & 92.6 & 90.0& 93.2 & 92.6 & 92.0 & 93.5 & 93.0 & 92.0 & 93.2 & 92.9 & 92.0 \\ 
			CM & 92.7 & 89.7 & 86.2 & 81.8 & 92.5 & 90.9 & 48.9 & 92.7 & 92.2 & 90.6 & 92.6 & 92.5 & 90.2 & 92.4 & 92.0 & 91.0 \\ 
			CMGT & 93.7 & 90.4 & 87.4 & 85.5 & 93.4 & 92.6 & 92.4 & 93.3 & 92.9 & 92.2 & 93.6 & 93.5 & 93.2 & 92.3 & 93.0 & 92.3 \\
			CT  & - & 90.6 & 86.9 & 78.6 & 93.1 & 92.4 & 71.5 & 93.3 & 92.7 & 68.1 & 93.3 & 92.5 & 91.2 & 93.4 & 92.8 & 92.3\\
			LS  & 93.6 & 90.3 & 86.4 & 77.6 & 93.1 & 92.5 & 89.3 & 93.2 & 92.6 & 92.1 & 93.2 & 92.9 & 91.8 & 93.2 & 92.8 & 92.3 \\
			NLS & 93.6 & 90.4 & 86.7 & 78.4 & 93.3 & 93.0 & 90.8 & 93.4 & 92.9 & 91.9 & 93.6 & 93.0 & 92.3 & 93.3 & 92.9 & 92.2 \\
			BTLS & 93.6 & 90.3 & 86.4 & 77.6 & 93.1 & 92.5 & 89.3 & 93.2 & 92.6 & 92.1 & 93.2 & 92.9 & 91.8 & 93.2 & 92.8 & 92.3 \\
			\bottomrule
	\end{tabular}}

\end{table*}

\section{Analysis by Acc on Different Set and State Transition}
\label{supplement:G}

\begin{figure}
	\centering
	\includegraphics[width=14cm]{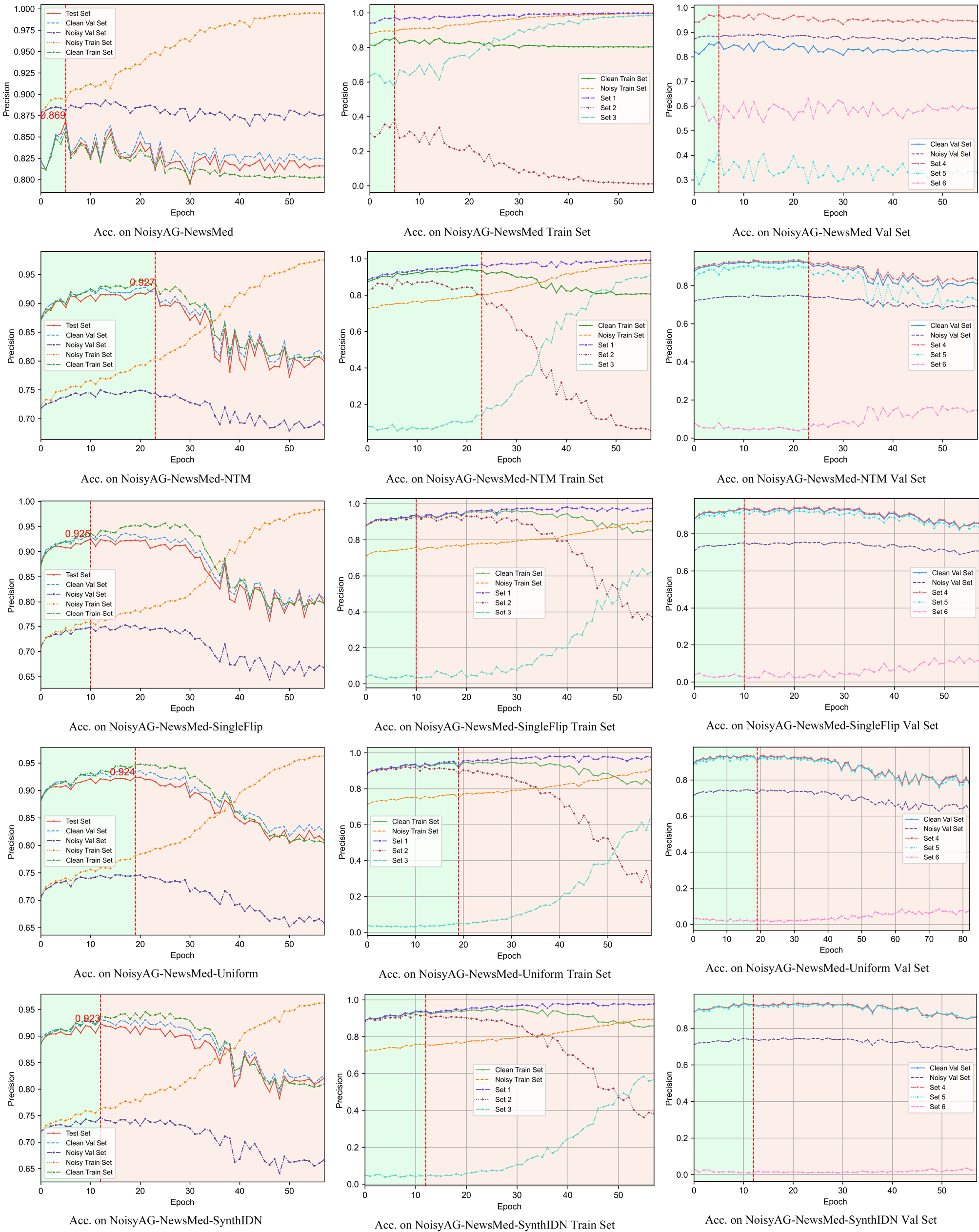}
	\caption{ Acc. on NoisyAG-NewsMed and Corresponding Synthetic Noise.  }
        \label{fig:medACCFinal_low}
\end{figure}

To gain a deeper understanding of the reasons behind the poor performance of classifiers on the NoisyAG-News dataset, we monitored the accuracy of classifiers on various datasets during training. By tracking accuracy on the test set, clean training set, noisy training set, clean validation set, and noisy validation set as depicted  in \ref{fig:medACCFinal_low} , we found that the performance on different synthetic noise datasets was quite similar, while the accuracy trends across the three NoisyAG-News subsets were also alike. All results indicate that models first fit the clean samples before fitting the noisy ones, as illustrated, causing the accuracy on the test set to initially increase and then decrease. The key difference is that on the NoisyAG-News dataset, the accuracy on the noisy training set approaches 1 more quickly, suggesting that the model can fit feature-dependent noisy labels faster. Models are generally less affected by class-dependent noise; when the noise rate increases from 0.1 to 0.38, the accuracy drops by no more than 5\%. However, for the NoisyAG-News dataset, the model's accuracy is strongly correlated with the noise rate and is significantly lower than the accuracy on synthetic noise datasets with corresponding noise rates. This indicates that current pre-trained large models are robust to class-dependent noise, while our proposed instance-dependent noisy dataset, NoisyAG-News, presents a new challenge.

To conduct a more detailed analysis of the performance of noisy labels and ground truth (GT) labels, we recorded the performance of Set1, Set2, Set3, Set4, Set5, and Set6 during training. The accuracy of the Clean Train Set is derived from the combination of Set1 and Set2, while the Noisy Train Set accuracy is derived from the combination of Set1 and Set3. Referring to Figure 3, we can similarly decompose the accuracy of the Test set, Clean Train Set, Noisy Train Set, Clean Val Set, and Noisy Val Set into the respective combinations of these sets.

As illustrated in \ref{fig:medACCFinal_low} , in the accuracy graph of synthetic noise, it can be observed that the model's accuracy on Set3 remains low in the early stages and only increases later. Conversely, the accuracy on Set2 is initially high but declines over time. This indicates that the model initially fits clean samples rather than noisy label data, and only later begins to fit the noisy data. Therefore, employing an early stopping strategy can yield a model with the best predictive accuracy, demonstrating that large models are robust to synthetic noise. Since the model fits the ground truth (GT) samples in the early stages, the noise rate primarily affects the number of GT samples fitted early on, making the test accuracy (TestAcc) insensitive to the noise rate. Additionally, by observing the accuracy on Set6, it is evident that even if the model fits Set3 data well, the accuracy on Set6 remains low. Furthermore, the lower the noise rate, the lower the accuracy on Set6, because synthetic noise is not feature-dependent.

In the accuracy chart for NoisyAG-News, the accuracy on the Noisy Train Set consistently improves. However, unlike synthesized noise, Set3 exhibits initially high accuracy that gradually stabilizes, while Set2 shows relatively low initial accuracy that declines over time. This indicates that the model initially fits the instance-related noisy labels, impacting its early performance on Set2.

During validation, because Set3 and Set6 are from the same distribution, unlike synthesized noise, Set6 achieves higher accuracy than Set5. This significantly lowers the accuracy on the Clean Validation and Test sets. By comparing the training and validation processes with synthetic noise and instance-related noise, we observe that the model is not robust to instance noise labels, easily fitting them. Additionally, since instance-related noise is feature-dependent, the model trained with it performs well on the Noisy Validation set. However, because Set6 and Set5 labels are contradictory, the accuracy on the Clean and Test sets is considerably lower.

\end{document}